\documentclass[journal]{IEEEtran}
\usepackage{balance}
\usepackage{amsfonts}
\usepackage{float}
\usepackage[pdftex]{graphicx}
\usepackage{subfigure}
\usepackage{epstopdf}
\usepackage[colorlinks,linkcolor=red]{hyperref}
\ifCLASSOPTIONcompsoc
\usepackage{booktabs}
  \usepackage[nocompress]{cite}
\else
  \usepackage{cite}
\fi
\usepackage{algorithm}
\usepackage{algorithmic}
\usepackage{amsmath}
\usepackage{makecell}
\usepackage{multirow}
\usepackage{color}
\usepackage{bm}

\usepackage{footnote}
\usepackage{threeparttable}

\begin{document}

\title{Credible Remote Sensing Scene Classification Using Evidential Fusion on Aerial-Ground Dual-view Images}
\newcommand{\w}{\textcolor{black}}
\newcommand{\s}{\textcolor{red}}
\author{Kun Zhao,~
        Qian Gao,~          
        Siyuan Hao,~
        Jie Sun,~
        Lijian Zhou* 
\thanks{
Corresponding author: Lijian Zhou.}
\thanks{Kun Zhao, Qian Gao, Siyuan Hao, Jie Sun and Lijian Zhou were with the School of Information and Control Engineering, Qingdao University of Technology, Qingdao 266520, China. 
E-mail: sterling1982@163.com, q17863936190@163.com, lemonbananan@163.com, sunjie1979@qut.edu.cn, zhoulijian@qut.edu.cn}
}
\markboth{Kun Zhao \MakeLowercase{\textit{et al.}}: Credible Remote Sensing Scene Classification Using Evidential Fusion on Aerial-Ground Dual-view Images}
{Kun Zhao \MakeLowercase{\textit{et al.}}: Credible Remote Sensing Scene Classification Using Evidential Fusion on Aerial-Ground Dual-view Images}
\maketitle

\begin{abstract}
Multi-view (multi-source, multi-modal and multi-perspective, etc.) data is increasingly used in remote sensing tasks because they can provide more useful information than single source data. However, data quality varies from view to view, limiting the potential benefits of multi-view data, which could be maximized by fusing them according to their credibility. Although recent deep learning based models are able to learn the weight of data adaptively, the lack of research on quantifying multi-view data credibility explicitly makes these models inexplicable, which affects their performance and flexibility in downstream remote sensing tasks. To fill this gap, in this paper, the theory of evidential deep learning is introduced to the aerial-ground dual-view remote sensing scene classification task to model the credibility of each view. Specifically, the theory of evidence is used to calculate an uncertainty value which describes the decision-making risk of each view. According to the uncertainty, a novel decision-level fusion strategy is proposed to make sure that the view with lower decision-making risk obtain more weight to make the classification more credible. Our approach achieved state-of-the-art performances on two classic public aerial-ground dual-view remote sensing image datasets, demonstrating the effectiveness of the proposed approach. 
\end{abstract}


\section{Introduction}
\label{sec: Introduction}
\IEEEPARstart{R}{emote} sensing scene classification, as a significant research area in remote sensing and satellite image analysis, is to classify scene images into discrete and meaningful land use and land cover categories according to different semantic features of remote sensing images. It is widely used in various practical applications, including geographic image retrieval~\cite {roy2020metric,wang2016three,tong2019exploiting}, natural disaster detection~\cite{martha2011segment,cheng2013automatic,stumpf2011object}, vegetation mapping~\cite{kim2009forest,capolupo2018novel,mishra2014mapping} and geo-spatial object detection~\cite{zhao2015dirichlet,yao2016semantic,wu2016hierarchical}.

In the past decades, great achievements have been made in designing efficient classifiers using a single data source, such as hyperspectral~\cite{liu2021multiscale}, multi-spectral~\cite{jiang2019multi}, synthetic aperture radar~\cite{chen2019new}, high resolution image~\cite{wang2018scene}. Although it is easier to obtain remote sensing data, remote sensing scene classification is still regarded as a challenging task~\cite{zhu2015land} when using only overhead images due to their lack of diverse detailed information. Fortunately, with the rise of various social media platforms (e.g. Weibo, Flickr, etc.) and mapping software (e.g. Google Maps, Bing Maps, etc.), it is becoming easier to collect geo-tagged data from various sources. Antoniou et al.~\cite{antoniou2016investigating} used geo-tagged social media images as input data and demonstrated that they provide vital information about land use and land cover. Pei et al. \cite{pei2014new} undertook land use classification at the mesh grid level using integrated data from mobile phones. Zhang et al.~\cite{zhang2017parcel} and Kang et al.~\cite{kang2018building} classified urban land functional areas using street view images. These studies show that the ground geo-tagged data can provide useful information for land use and land cover classification.

Research on fusing overhead images with other geo-tagged data for remote sensing scene classification has attracted more and more attention. Hu et al.~\cite{hu2016mapping} and Liu et al.~\cite{liu2017classifying} combined satellite images with points of interest (POIs) and other social media data to classify urban parcels, demonstrating the potential of social media data to enhance scene classification. Jia et al.~\cite{jia2018urban} adopted a support vector machine to classify the features extracted from remote sensing images and mobile phone location data separately, fusing the two classification results using decision-level fusion for the classification of urban land use. Tu et al.~\cite{tu2018portraying} proposed a data fusion framework to combine remote sensing imagery and human sensing data, using a hierarchal clustering method to identify urban functional zones. Hong et al.~\cite{hong2019cospace} proposed common subspace learning (CoSpace) on hyperspectral-multispectral correspondences, which translates the remote sensing scene classification problem into a cross-modal learning problem.

\begin{figure*}[ht]
\centering
\includegraphics[width=16cm]{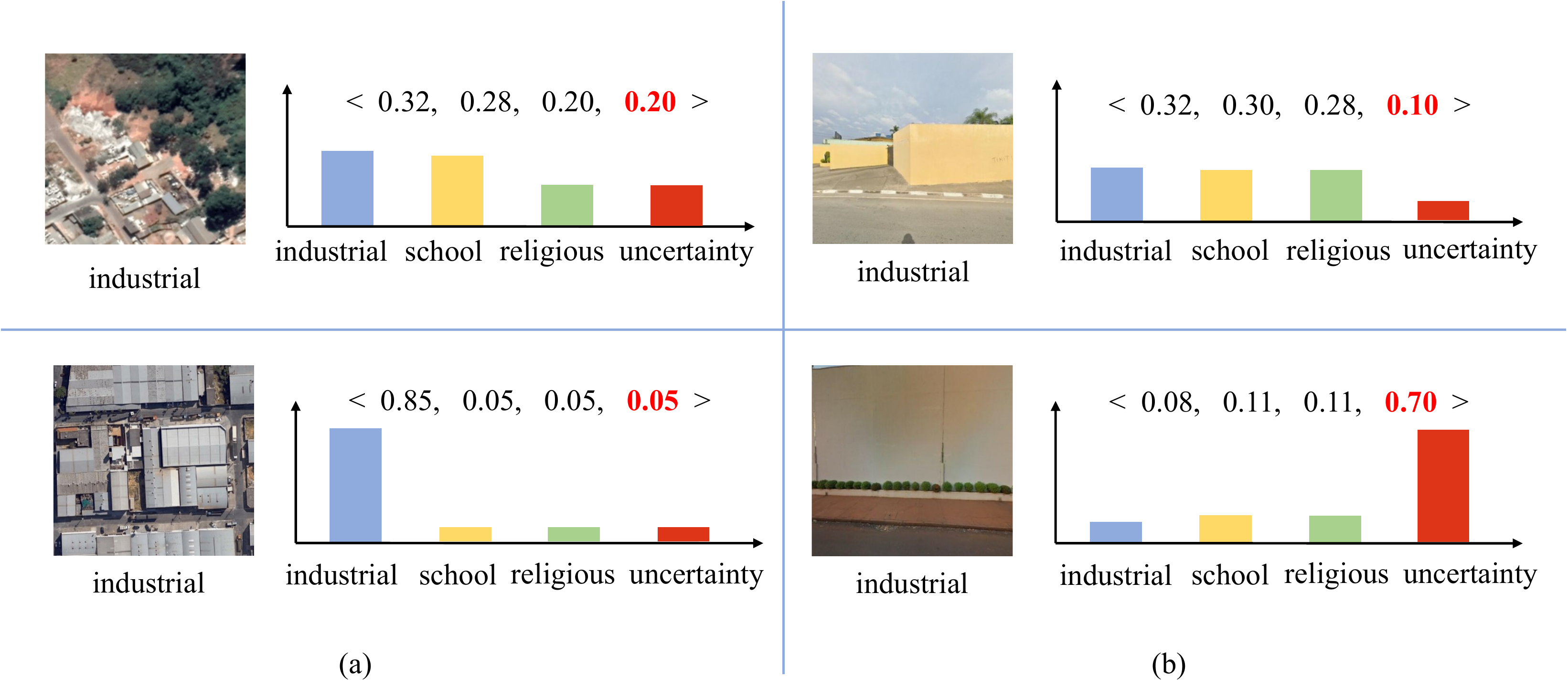}
\caption{Which image is more credible? Different kinds of sample uncertainty in (a) aerial images and (b) ground images. The first three bins in each histogram are the possible probabilities of each class. The last red bin in each histogram reference to the probable uncertainty values.}
\label{fig:motivation}
\end{figure*}

Among many multi-source data for remote sensing scence classification, aerial-ground dual-view images are frequently
used due to their widespread geographic availability and easy of access~\cite{li2017building}. Cao et al.~\cite{cao2018integrating} extracted semantic features from sparsely distributed street view images to match the spatial resolution of the aerial images, which are then fused together through a deep neural network for classifying land use and land cover. Srivastava et al.~\cite{srivastava2019understanding} used end-to-end learning of single-mode feature extraction and fusion to predict land cover labels from remote sensing images, street view images and annotations from open street maps. It is better to use decision-level fusion than direct data-level fusion for scene classification~\cite{machado2020airound}. Due to aerial-ground dual-view images are represented by heterogeneous features with different distributions, Deng et al.~\cite{deng2018semi} proposed a semi-supervised manifold-regularized multiple-kernel-learning (SMRMKL) algorithm for solving these problems.

However, the use of multi-source data also brings about an increase in sample uncertainty~\cite{han2021trusted}. In particular, as a special type of remote sensing multi-modal data, the problem of sample uncertainty for aerial-ground dual-view images is more serious, due to the inter-class similarity and intra-class diversity of aerial images and the poor quality (e.g. severe occlusion, indoor perspective, etc.) of ground images. For example, Fig.\ref{fig:motivation}(a) shows two aerial images of industrial area. The bottom image is obviously more credible because it clearly depicts an industrial area. Well, the top image should have a higher degree of uncertainty because it also appears to be a school area. The quantification of sample uncertainty (red bins) can help to describe the inter-class similarity and intra-class diversity which are common in aerial images, contributing to better classification. Fig.\ref{fig:motivation}(b) shows another usual situation. In this case, the top ground view image depicts a scene that can be seemed as any of three classes, while the bottom one has no useful information at all. It is clear that the top image contains more useful information and thus is more credible. However, without quantification of sample uncertainty (red bins), these two images will play the same role in scene classification. Unfortunately, so far there is not enough research on sample uncertainty in multi-view remote sensing data fusion.

From the idea of explicitly modeling the sample uncertainty, a novel fusion approach is proposed in this paper based on evidential deep learning~\cite{sensoy2018evidential} for remote sensing scence classification on aerial-ground dual-view images. The primary contributions of this paper are as follows.

\begin{itemize}
\item 
A \textbf{Theory of Evidence Based Sample Uncertainty Quantification (TEB-SUQ)} approach is used in both views of aerial and ground images to measure the decision risk during their fusion.

\item 
An \textbf{Evidential Fusion} strategy is proposed to fuse aerial-ground dual-view images in decision-level for remote sensing scene classification. Unlike other existing decision-level fusion networks, the proposed strategy focuses the results not only on the classification probability but also on the decision risk of each perspective. As a result, the final result will depend more on the view with lower decision risk.

\item 
A more concise loss function, namely \textbf{Reciprocal Loss} is designed to simultaneously constrain the uncertainty of individual view and the uncertainty of their fusion. It can be used not only to train an end-to-end aerial-ground dual-view remote sensing scene classification model, but also to train a fusion classifier without feature extraction.
\end{itemize}

\section{Related Work}
\subsection{Multi-view Data Fusion in Remote Sensing}
\label{subsec: Multi-view Data Fusion in Remote Sensing}

The goal of multi-view learning is to utilize or correlate data from various views (e.g., various sources, modalities or perspectives) for model building \cite{baltruvsaitis2018multimodal}. The fusion approaches for multi-view learning can be implemented at three levels~\cite{ramachandram2017deep}, namely the \textbf{data-level}, the \textbf{feature-level} and the~\textbf{decision-level}. The~\textbf{data-level} fusion strategy usually fuses raw or pre-processed data from several data sources~\cite{tu2004fast} or multi-resolution images~\cite{amolins2007wavelet}, etc. Multi-scale wavelets~\cite{nunez1999multiresolution}, bi-dimensional empirical mode decomposition\cite{liu2007bidimensional} and Laplacian pyramids~\cite{aiazzi2002context} are used to fuse the data of multi-spectral image. \textbf{Feature-level} fusion, on the other hand, combines multiple intermediate features extracted from multi-view data. The fused features were then used in downstream tasks, such as being classified into land use and land cover through object-oriented multi-scale segmentation \cite{zhang2020novel}. Wang et al.~\cite{wang2022multi} introduced a multi-layer attention fusion network which hierarchically fused and classified the features of hyperspectral images and laser ranging data (LiDAR). \textbf{Decision-level fusion} adopts different fusion rules to aggregate predictions from multiple classifiers~\cite{ye2012robust, gunes2005affect}, each of which is obtained from a separate model. Yu et al.~\cite{yu2018aerial} input the aerial image and the two patches extracted from the image into three convolutional neural networks with different receptive fields, and then use the probability fusion model for the final classification. Yang et al.~\cite{yang2018dropband} proposed a scene classification approach for very high resolution remote sensing images by discarding some spectra of the original image, inputting them into convolutional neural networks (CNNs) and combining their output predictions. Zhang et al.~\cite{zhang2015scene} introduced a gradient enhanced random convolution network (GBRCN) structure that can successfully combine numerous deep neural networks for scene classification in decision-level fusion. Because the features of each view are learned individually and do not affect each other, the decision-level fusion is the most flexible of the three levels. It is very suitable for the remote sensing scene classification task of the aerial-ground dual-view images due to the difference in the structure of data in different views can be ignored. However, the quality of the fusion depends more on the credibility of the decision results of each view. Therefore, it is particularly important to quantify the uncertainty of the predictions of each view.

\subsection{Aerial-Ground Dual-view Image Fusion in Remote Sensing}

Unlike other commonly used multi-view remote sensing images (e.g. Satellite and LiDAR, multi-spectral images, etc. ), the difference between the matched aerial image and ground image are too huge to be fused at data-level directly. Thus, fusion methods at feature-level and decision-level are often used. Aerial-ground dual-view image fusion was initially used in image geo-localization. Lin et al.~\cite{lin2013cross} matched high resolution orthophoto (HRO) from Bing Maps with street view images from Panoramio by using four handcrafted features and adding land cover features as the third modality. To extend their approach, they used a Siamese-like CNN to learn deep features between Google Street View (GSV) images and 45-degree oblique aerial images~\cite{lin2015learning}. Workman et al.~\cite{workman2017unified} fused aerial images and GSV images by an end-to-end deep network which outputs a pixel-level labeling of aerial images for three different classification problems: land use, building function and building age. Zhang et al.~\cite{zhang2017parcel} fused airborne light detection and ranging (LiDAR) data, HRO and GSV images for land use classification. In their study, thirteen parcel features were chosen as input variables in a random forest classifier. Cao et al.~\cite{cao2018integrating} used images from Bing Maps and GSV for land use segmentation with a two-stream encoder and one decoder architecture which evolved from SegNet~\cite{badrinarayanan2017segnet}. Hoffmann et al.~\cite{hoffmann2019model} used a two-stream CNN model for building functions classification. They predicted four class labels namely commercial, residential, public and industrial for overhead images by fusing deep features of aerial images and ground view images. Their model increased the precision score a lot with a decision-level fusion strategy. Srivastava et al.~\cite{srivastava2019understanding} extend their early work~\cite{srivastava2018multilabel} to a multi-modal strategy by leveraging the complementarity of aerial-ground dual-view. They deal with the situation of missing aerial images by using canonical correlation analysis (CCA) based on their two-stream CNN model.

\subsection{Uncertainty Estimation}

In order to describe the results more comprehensively, we are pursuing not only an increase in the classification accuracy, but also a risk evaluation of the predictions. Uncertainty estimation is devoted to output the uncertainty of predictions and reducing the overconfidence. A great deal of researches have been done on uncertainty estimation. Jiang et al.~\cite{jiang2018trust} suggested the term of “trust score” as a measurement of confidence using a confidence criterion which is the ratio between the distance from the sample to the nearest class and the distance to the predicted class. However, it is computationally intensive. Another drawback is that the local distance calculation~\cite{beyer1999nearest} does not make sense in high-dimensional spaces and may have a negative impact on the result. Recently, there has been a significant amount of attention on Bayesian methods for uncertainty estimation in neural networks. By distributing over the model parameters and marginalizing these parameters, Bayesian neural networks (BNNs) \cite{blundell2015weight} modeled uncertainty regarding a predictive distribution. True class probability (TCP)~\cite{corbiere2019addressing} proposed an additional module to derive the uncertainty of the output, which is a new criterion that also provides great help in predicting failure. Subjective logic was used in evidential deep learning (EDL)~\cite{sensoy2018evidential} to model the uncertainty of the output probability. The Dirichlet distribution was placed on the class probabilities. Deep ensembles~\cite{lakshminarayanan2017simple} captures “model uncertainty” by averaging the predictions of multiple models which are consistent with the training data. Deterministic uncertainty quantification (DUQ)~\cite{van2020uncertainty} calculated radial Basis function (RBF) distances by the mutual placement of mass centers to convey uncertainty. A deep belief network based on Gaussian process mappings, namely deep Gaussian process~\cite{damianou2013deep} modeled sample uncertainty with non-parametric kernel functions. The uncertainty-aware attention (UA) mechanism~\cite{heo2018uncertainty} is proposed to capture heteroscedastic uncertainty, or the instance-specific uncertainty, which in turn yields more accurate calibration of prediction uncertainty. With increasing amounts of research on uncertainty, it is possible to observe a reduction in uncertainty that can make the decisions made by the network more deterministic~\cite{blundell2015weight}. Unfortunately, all of the methods presented above focused on estimating the uncertainty on single-view data, despite the fact that fusing multi-views using uncertainty can improve performance and credibility. 

\begin{figure*}[ht]
\centering
\includegraphics[width=16cm]{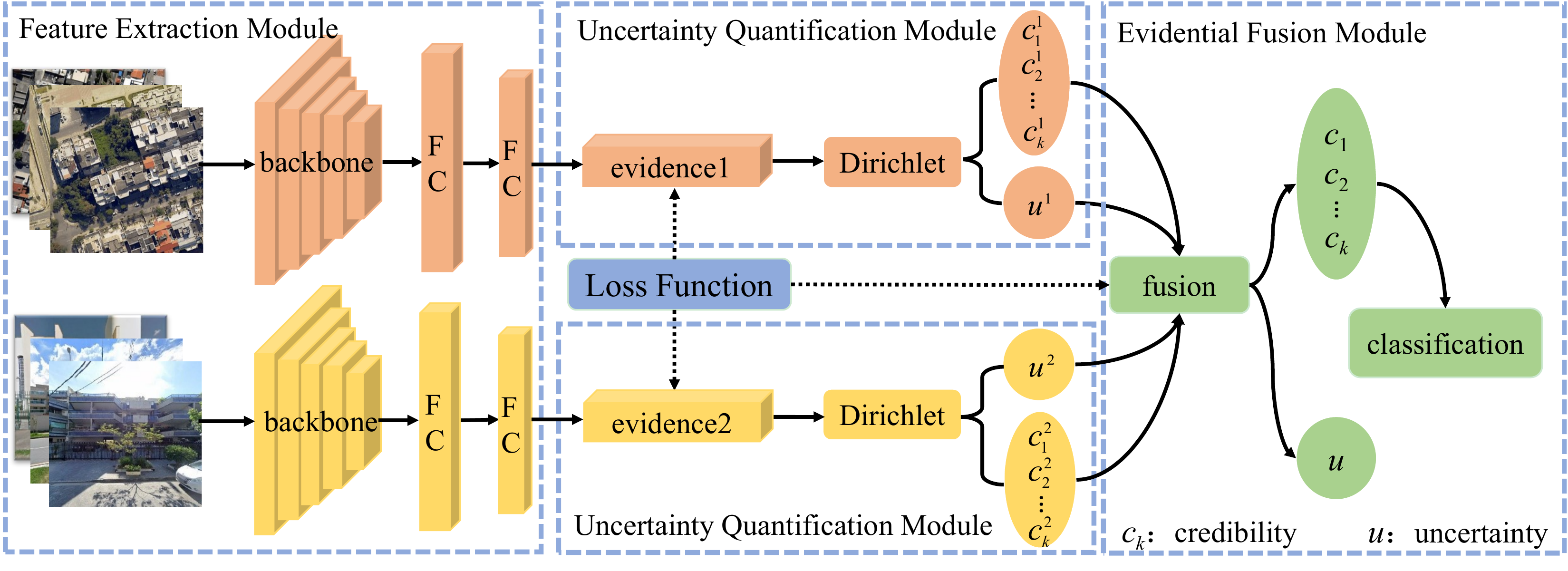}
\caption{The overall framework of proposed Evidential Fusion Network (EFN) on aerial-ground dual-view images.}
\label{fig:pipeline}
\end{figure*}

\section{Methodology}
In this section, the \textbf{Evidential Fusion Network (EFN)} for remote sensing scene classification on aerial-ground dual-view images is introduced. First, the overall network architecture is introduced briefly. Second, we detail how to obtain sample uncertainty with evidential deep learning and how to perform the evidential fusion. Finally, the proposed \textbf{Reciprocal Loss} used to train the network is described.

\subsection{Overview}
\label{subsec: Overview}
Recent works~\cite{machado2020airound,srivastava2019understanding} show that decision-level fusion using a deep learning framework usually performs best in the task of remote sensing scene classification on aerial-ground dual-view images. However, due to the lack of measurement of their credibility, multi-view decisions (namely, the inferred results of the input samples) play equally important roles during the fusion, despite the fact that their samples have unequal uncertainty. On the one hand, although the qualities of aerial view images are relatively higher, the issue of intra-class diversity and inter-class similarity is more prominent. On the other hand, despite providing more distinguishable details, ground view images often suffer from poor qualities due to the problems such as occlusion, large variations in perspective and poor positioning accuracy. Without taking sample uncertainty into account, the fusion will be misled by a large number of “over-confidence”. The proposed evidential fusion network (EFN) can assess the decision risk of each view by quantifying the sample uncertainty explicitly, and then assign different weight to the decisions base on their risks when fusing them. As a result, the final classification relies more on the view with lower decision risk therefore become more credible. The overall framework of EFN is shown in Fig.\ref{fig:pipeline}. It can be further divided into three modules: the feature extraction module (FEM), the uncertainty quantification module (UQM) and the evidential fusion module (EFM).

In each view, the trained backbone model is used to extract one-dimensional feature vector which is then sent into the fully connected layer with non-negative activation function to obtain the “evidence” for the uncertainty computation. Then, the evidence is individually mapped to the concentration parameters of the Dirichlet distribution which is used to obtain the credibility and the uncertainty (see Section~\ref{subsec: Uncertainty Estimation} for more details). Finally, a weighted decision-level fusion is adopted to fuse the credibility and the uncertainty of aerial-ground dual-view images (see Section~\ref{subsec: Evidential Fusion} for more details). It makes full use of the uncertain information of each view, which can reduce the weight of the views with higher decision risk when participating in the final decision. 

The proposed FEM is divided into two subnets, each with a backbone and two extra fully connected layers. The backbones of two the subnets could be the same or different because the proposed approach is independent of the backbones. It is this flexibility that makes the proposed approach applicable to any network structure. Finally, the softmax operator at the end of the last fully connected layer is replaced with a non-negative activation function, such as ReLU. This simple replacement operation makes the proposed approach extremely portable.

\subsection{Uncertainty Estimation Based on Evidential Learning}
\label{subsec: Uncertainty Estimation}

\begin{figure}[ht]
\centering
\includegraphics[width=\linewidth]{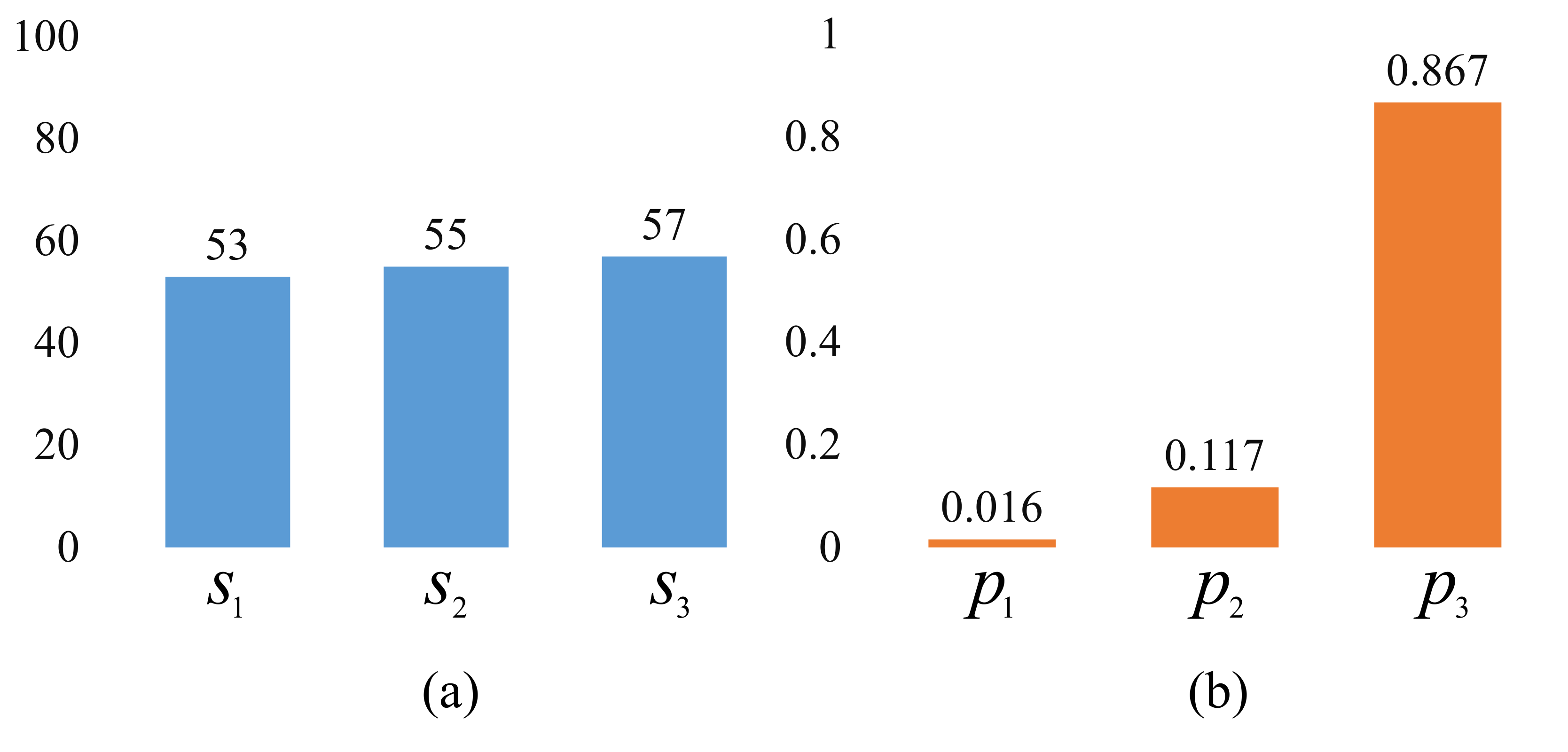}
\caption{The “over-confidence” caused by softmax: (a) are the probable scores of three categories after feature extraction and (b) are their corresponding probabilities mapped by softmax.}
\label{fig:over-confidence}
\end{figure}

\begin{figure}[ht]
\centering
\includegraphics[width=\linewidth]{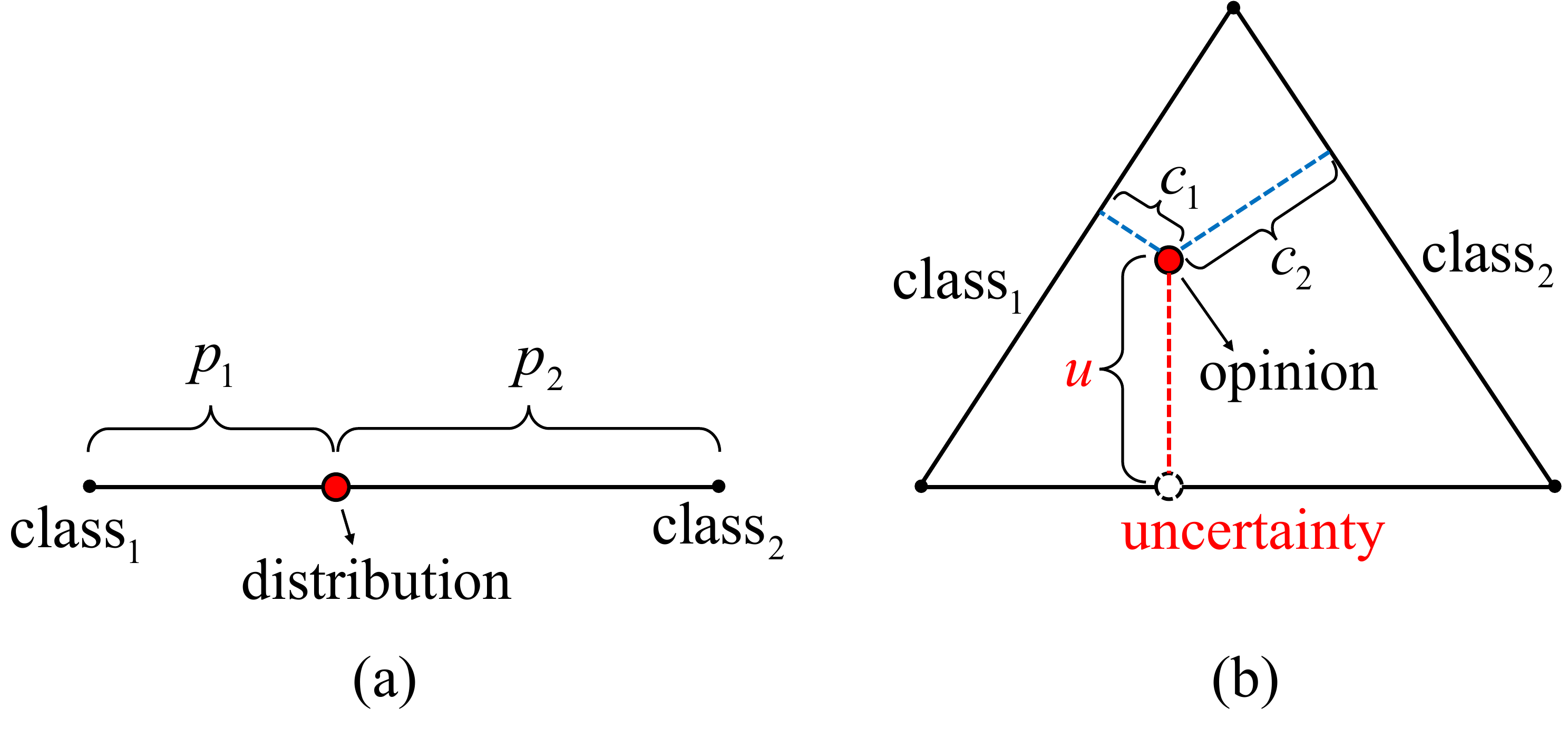}
\caption{The case of a binary classification problem: (a) is the 
 class probability distribution of a sample generated by softmax; (b) is the “opinion” of the same sample generated by the an uncertainty estimation operation where $u$ is the uncertainty and $c_i$ is the credibility of class $i$.}
\label{fig:binary-classification}
\end{figure}

\begin{figure*}[ht]
\centering
\includegraphics[width=14cm]{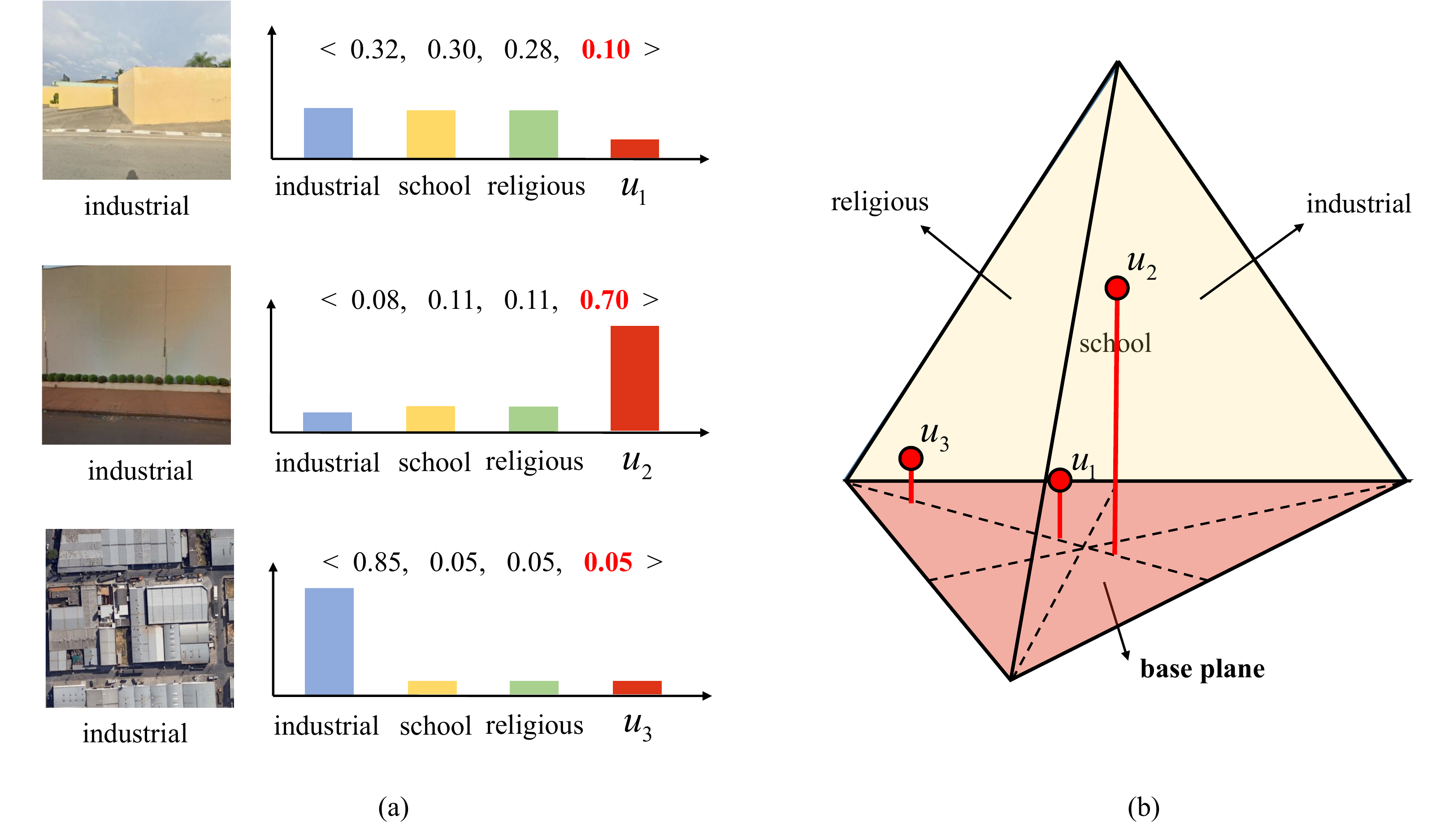}
\caption{A multi-classification case of the proposed uncertainty estimation operation: (a) shows the “opinions” of three samples (two ground view image and one aerial image); (b) is the Opinion tetrahedron with example opinions.}
\label{fig:multi-classification}
\end{figure*}

As is known to all, the use of the softmax operator to transform the output continuous activations into discrete class probabilities is the gold standard for classification networks. However, recent studies show that softmax operator has many shortcomings. Firstly, it may lead to the problem of “over-confidence”~\cite{moon2020confidence}, namely it will compel the uncertain samples to be classified into a certain category by magnifying the differences between predicted scores in different categories. Taking Fig.~\ref{fig:over-confidence} as an example, the scores of the three categories in Fig.~\ref{fig:over-confidence}(a) are very close, indicating that the model is unsure about the category of the input sample. However, after the softmax mapping in Fig.~\ref{fig:over-confidence}(b), the difference between the probabilities is greatly increased, giving the model the false impression of being very certain about the results.

Secondly, softmax can only provide a point estimate for the category probability of a sample without providing the associated uncertainty, which often leads to unreliable conclusions~\cite{van2020uncertainty}. When there is no relevant feature in the input image, it will be induced or even forced to set a result in the final prediction without any evidence to support it~\cite{neumann2018relaxed}. Fig.~\ref{fig:binary-classification} shows the difference between the softmax and an uncertainty estimation operation on a case of a binary classification problem. In Fig.~\ref{fig:binary-classification}(a), a class probability distribution is generated by softmax, where $p_i$ is the probability of class $i$, and we have $p_1+p_2=1$ and $p_1<p_2$, which means that the classifier is more inclined to classify the sample into class 2. However, we have no idea whether this prediction is credible. To assess the credibility of a sample, the quantification of uncertainty needs to be explicitly modeled.

In order to address the drawbacks of softmax for scene classification, a non-negative activation function (such as ReLU) which replaces the softmax operator is used to output the “evidence” of each class for samples. Based on these evidences, the sample uncertainty can be described by Dempster-Shafer Theory (DST) of evidence~\cite{yager2008classic}, which allows to explicitly express “ignorance”, i.e. the lack of evidence about the truth of a sample by taking the “base plane” (see Fig.~\ref{fig:multi-classification}(b)) into consideration when building the “opinion space”, and further assigning the overall credibility of the sample to the possible class labels. In Fig.~\ref{fig:binary-classification}(b), an “opinion” (the red point in the equilateral triangle) of the same sample is generated by an uncertainty estimation operation, where $c_i$ is the credibility of class $i$ which is equal to the distance from the “opinion” to the side representing class $i$, and $u$ is the sample uncertainty which is equal to the distance from the “opinion” to the base. It is well known that the sum of the distances from any point in an equilateral triangle to its three sides is equal to the height of the equilateral triangle. When setting the height of the equilateral triangle to 1, We have $c_1+c_2+u=1$ and $p_1<p_2<u$, which means that the classifier is more likely to find the sample unreliable than to make a binary decision. 

Fig.~\ref{fig:multi-classification} shows examples with three classes, which have been shown in Fig.~\ref{fig:motivation}. In this case, the opinion equilateral triangle in Fig.~\ref{fig:binary-classification} rises its dimension to become a opinion tetrahedron~\cite{josang2016subjective} where each class is denoted as a side plane. The credibility of each class is equal to the distance from the opinion point to the corresponding side plane. And the vertical elevation of the opinion point represents the sample uncertainty. 

In a more general sense, in a space with $K$ mutually exclusive singletons such as class labels, DST can provide a credibility for each singleton and an overall uncertainty. The prediction of a sample is often referred to as an “opinion”. The $K$ side planes are the possible classes whose credibility values are the distances between them to the opinion point. The uncertainty of an opinion is interpreted as a measure of the truth of the sample. These $K$ credibility values $c_{k}$ and the uncertainty $u$ are both non-negative and sum to one, that is,

\begin{equation}
 u+\sum_{k=1}^K c_{k} = 1,
\label{eq:u+c}
\end{equation}
where $u \geq 0$ and $c_{k}\geq 0$ for $k=1,2,\cdots,K$, denote the overall uncertainty and the credibility of $k$-th class respectively. For each view, the credibility of each class can be calculated using the evidence of each class. The non-negative evidence set of a sample can be denoted as $\mathbf{e}=[e_{1},e_{2},\cdots,e_{K}]$. Let $e_{k}$ be the evidence of $k$-th class of the sample, therefore, the credibility $c_{k}$ and the uncertainty $u$ can be calculated as

\begin{equation}
 c_{k} = \frac{e_{k}}{\sum_{k=1}^K (e_{k}+1)},
\label{eq:ck}
\end{equation}

\begin{equation}
 u = \frac{K}{\sum_{k=1}^K (e_{k}+1)}.
\label{eq:u}
\end{equation}

From the above equation, it can be seen that the sample uncertainty is inversely proportional to the overall evidence. When no evidence is learned, the credibility of each class is 0, and the sample uncertainty is 1. As more and more evidence is available for the classifications, the credibility increases while the sample uncertainty decreases.

The notion of DST could be further formalized as a Dirichlet distribution~\cite{josang2016subjective} which is based on a probability “range” rather than individual probability values. It is this “range” that plays a central role in quantifying uncertainty: when the probability range of a certain category is too wide, it means that the classifier is not sure about the opinion; on the contrary, when it shrinks to a discrete probability value, it means that the classifier is very sure about the prediction. In other words, by using the Dirichlet distribution (also known as the “distribution of distributions”), the sample uncertainty could be interpreted as a second-order probability of a first-order probability. And this “range” can be precisely described by the concentration parameter of the Dirichlet distribution. In contrast, the output of a standard neural network classifier is a discrete probability assignment of the possible classes for each sample, which is “so sure” that it would lead to the over-confidence.

To be exact, in order to calculate the sample uncertainty by DST, each opinion will correspond to a Dirichlet distribution with concentration parameter
\begin{equation}
 \alpha_{k}=e_{k}+1.
\label{eq:alpha}
\end{equation}
That is, the credibility and uncertainty can be easily obtained from the corresponding Dirichlet distribution using the following equations respectively:
\begin{equation}
 c_{k}=\frac{\alpha_{k}-1}{\alpha_{0}}, 
\label{eq:ck-new}
\end{equation}

\begin{equation}
 u=\frac{K}{\alpha_{0}}, 
\label{eq:u-new}
\end{equation}

\begin{equation}
 \alpha_{0}=\sum_{k=1}^K\alpha_{k}.
\label{eq:alpha-0}
\end{equation}
For a $K$-classification problem, the concentration parameter $\bm{\alpha}=[\alpha_{1},\alpha_{2},\cdots,\alpha_{K}]$, and the Dirichlet distribution is given by
\begin{equation}
 D(\mathbf{p}|\bm{\alpha})=
 \begin{cases}
  \frac{1}{B(\bm{\alpha})}\prod_{k=1}^K p_{k}^{\alpha_{k}-1}  & \mathbf{p}\in S_{K}\\
  0                       & otherwise
 \end{cases}
,
\label{eq:Dirichlet-distribution}
\end{equation}
where $B(\bm{\alpha}) $ is a $K$-dimensional multivariate beta function and $\mathcal{S}_{K}$ is the $K$-dimensional unit simplex,
\begin{equation}
 \mathcal{S}_{K} =  \lbrace \mathbf{p}|\sum_{k=1}^K p_{k} = 1\quad and\quad0 \leq p_{1},p_{2},\cdots,p_{K} \leq1 \rbrace.
\label{eq:S}
\end{equation}


When a scene image is input, the corresponding concentration parameter of the Dirichlet distribution will be increase if it can be observed to be similar to one of the $K$ classes. The Dirichlet distribution is then updated with new samples. Therefore, the concentration parameter of the Dirichlet distribution is associated with the evidence of each class. Up to now, the \textbf{TEB-SUQ} in this section is ready to be used on both views of aerial and ground images to measure the decision risk during their fusion.

\begin{figure*}[ht]
\centering
\includegraphics[width=\linewidth]{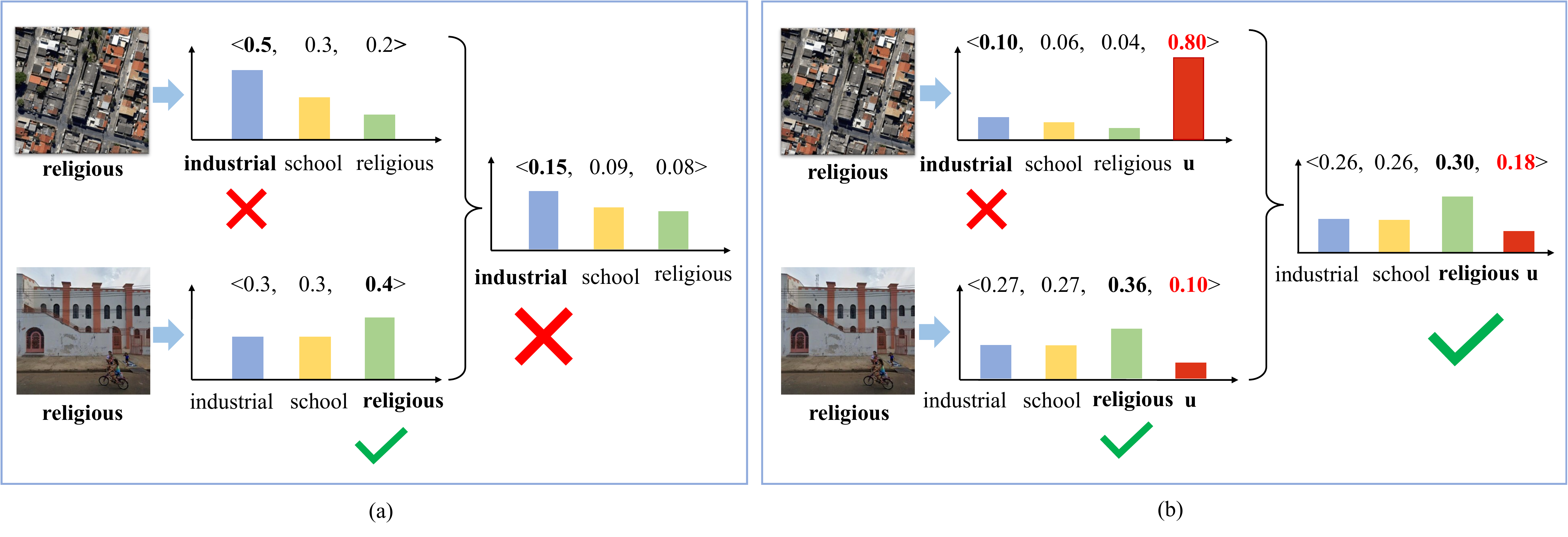}
\caption{Different predictions are made when using different fusion strategies on aerial-ground dual-view decisions: (a) using a multiplication fusion on the softmax outputs of both views; and (b) using the proposed evidential fusion on the UQM outputting opinions. All predictions are the class label with the highest score, whether in a single view or after fusion.}
\label{fig:fusion-example}
\end{figure*}

\subsection{Evidential Fusion}
\label{subsec: Evidential Fusion}
For scene classification, the fusion of dual-view images can indeed make use of the complementary information of different views. However, due to the problems of aerial and ground view images mentioned in Section~\ref{sec: Introduction} (see Fig.~\ref{fig:motivation}), directly fusing the softmax outputs of dual-view will ignore the decision risks of the views, thus obtaining unreliable results. Uncertainty estimation is the key to quantify the credibility of an opinion, which has been mentioned in Section~\ref{subsec: Uncertainty Estimation}. By taking advantage of the sample uncertainty in both views, potential benefits of dual-view images will be maximized and the classification result will become more reliable.

In order to reduce the weight of the view with higher decision risk in the final classification, a novel decision-level fusion strategy, namely \textbf{Evidential Fusion} is proposed based on Eq.~\ref{eq:ck} and Eq.~\ref{eq:u} in this section, which uses the sample uncertainty of each view as its decision risk to allocate the weight in the final prediction.

To be more precise, the opinions $\mathcal{O}^1=\lbrace c_{1}^1, c_{2}^1, \dots, c_{K}^1,u^1 \rbrace$ and $\mathcal{O}^2=\lbrace c_{1}^2, c_{2}^2, \dots, c_{K}^2,u^2 \rbrace$ of the two views are obtained after the UQM, where the superscripts denote the number of different views. The final decision opinions $\mathcal{O}=\lbrace c_{1},c_{2},\dots,c_{K},u \rbrace$ is then calculated as follows:
\begin{equation}
 c_{k}={\frac{1}{\lambda}}[c_{k}^1c_{k}^2+(1-u^1)c_{k}^1+(1-u^2)c_{k}^2],
\label{eq:ck-fusion}
\end{equation}

\begin{equation}
 u={\frac{1}{\lambda}}u^1 u^2,
\label{eq:u-fusion}
\end{equation}

\begin{equation}
 \lambda=u^1 u^2+(1-u^1)^2+(1-u^2)^2+\sum_{k=1}^K c_{k}^1c_{k}^2,
\label{eq:lambda}
\end{equation}
where $\lambda$ is scale factor to perform the normalization and to ensures that Eq.~\ref{eq:u+c} still holds after fusion.

The final decision-making opinions are formed by the fusion of opinions from two views. The fusion strategy is designed to cover three general cases. 

\begin{itemize}
\item 
When both views have low decision risk (both $ u^1 $ and $ u^2 $ are small), that is, both views have high credibility, the final prediction should have high credibility ( $ c $ will be large). 

\item 
When one of the two views has high decision risk and the other view has low decision risk (only $ u^1 $ or $ u^2 $ is large), the final prediction should rely more on the one with lower decision risk. 

\item 
When both views have high decision risk (both $ u^1 $ and $ u^2 $ are large), the credibility of both views is low, the final prediction should have low credibility ($ c $ will be small).
\end{itemize}

After obtaining the final decision opinion $\mathcal{O}$ from the fusion of two views, the fused evidence $ e_{k} $ of the $ k $-th class can be calculated according to the following equation:

\begin{equation}
 e_{k}= \frac{K\cdot c_{k}}{u},
\label{eq:ek-update}
\end{equation}
and the concentration parameters $ \alpha_{k} $ of the Dirichlet distribution of the $ k $-th class can be updated by Eq.~\ref{eq:alpha} to calculate the loss during training step and the category with the largest $ e_{k} $ is the final predicted label during test step.

An example is given to demonstrate how the the proposed evidential fusion works in Fig.~\ref{fig:fusion-example}. A multiplication fusion on the softmax outputs of aerial-ground dual-view images is shown in Fig.~\ref{fig:fusion-example}(a). The final prediction depends equally on the decisions of both views. The neglect of decision risk leads to wrong classification results. The proposed evidential fusion on the UQM outputting opinions is shown in Fig.~\ref{fig:fusion-example}(b). Based on the uncertainty (decision risk) of each view calculated by the TEB-SUQ, the final prediction depends more on the decision of the view with less risk, thus obtaining the correct classification result. Or it can be said that the proposed approach obtains a more credible classification result.

\subsection{Loss Function}
\label{subsec: Loss Function}
As has been mentioned in Section~\ref{subsec: Overview}, in the FEM, the softmax layer was replaced by a non-negative output activation layer (e.g. ReLU, softplus, etc.) to provide evidence vectors for the Dirichlet distribution prediction. To train the proposed model, the following loss function is designed and then adopted on each view:

\begin{equation}
 L(\bm{\alpha})=\sum_{i=1}^N(L_{pc}(\bm{\alpha}_{i})+L_{nc}(\bm{\alpha}_{i})),
\label{eq:overall-loss}
\end{equation}

\begin{equation}
 L_{pc}(\bm{\alpha}_{i})=\sum_{k=1}^K y_{ik}\left[\psi(\alpha_{i0})-\psi(\alpha_{ik})\right],
\label{eq:pc-loss}
\end{equation}

\begin{equation}
 L_{nc}(\bm{\alpha}_{i}) =\sum_{k=1}^K (1-y_{ik}) \frac{1}{\psi(\alpha_{i0})-\psi(\alpha_{ik})},
\label{eq:nc-loss}
\end{equation}
where $ L_{pc}(\bm{\alpha}_{i}) $ and $ L_{nc}(\bm{\alpha}_{i})$ are positive-class loss and negative-class loss of the $i$-th sample respectively, $ \psi(\cdot)$ is the digamma function which is monotonically increasing in $(0,+\infty)$, $\alpha_{i0}$ is consistent with Eq.~\ref{eq:alpha-0}, $K$ is the number of classes and $N$ is the number of samples. 

As shown in Eq.~\ref{eq:overall-loss}, the proposed loss function clearly separates the penalty terms of the positive and negative classes, making it easy to interpret. Furthermore, Eq.~\ref{eq:pc-loss} and Eq.~\ref{eq:nc-loss} show that the positive-class loss and the negative-class loss are reciprocal. Thus, the proposed loss function is called “Reciprocal Loss” in this paper. More specifically, since $\alpha_{ik}=e_{ik}+1$ and $e_{ik}>0$, so $\psi(\alpha_{ik})$ is monotonically increasing with $\alpha_{ik}$, while $L_{pc}(\bm{\alpha}_{i})$ is monotonically decreasing with $\alpha_{ik}$, which ensures that the positive class of each sample generates more evidence. On the contrary, $L_{nc}(\bm{\alpha}_{i})$ is monotonically increasing with $\alpha_{ik}$ to ensure that negative classes of each sample generate less evidence. Next, taking $L_{pc}(\bm{\alpha}_{i})$ as an example, how the proposed Reciprocal Loss is related to the Dirichlet distribution will be derived.

\begin{figure}[ht]
\centering
\includegraphics[width=9cm]{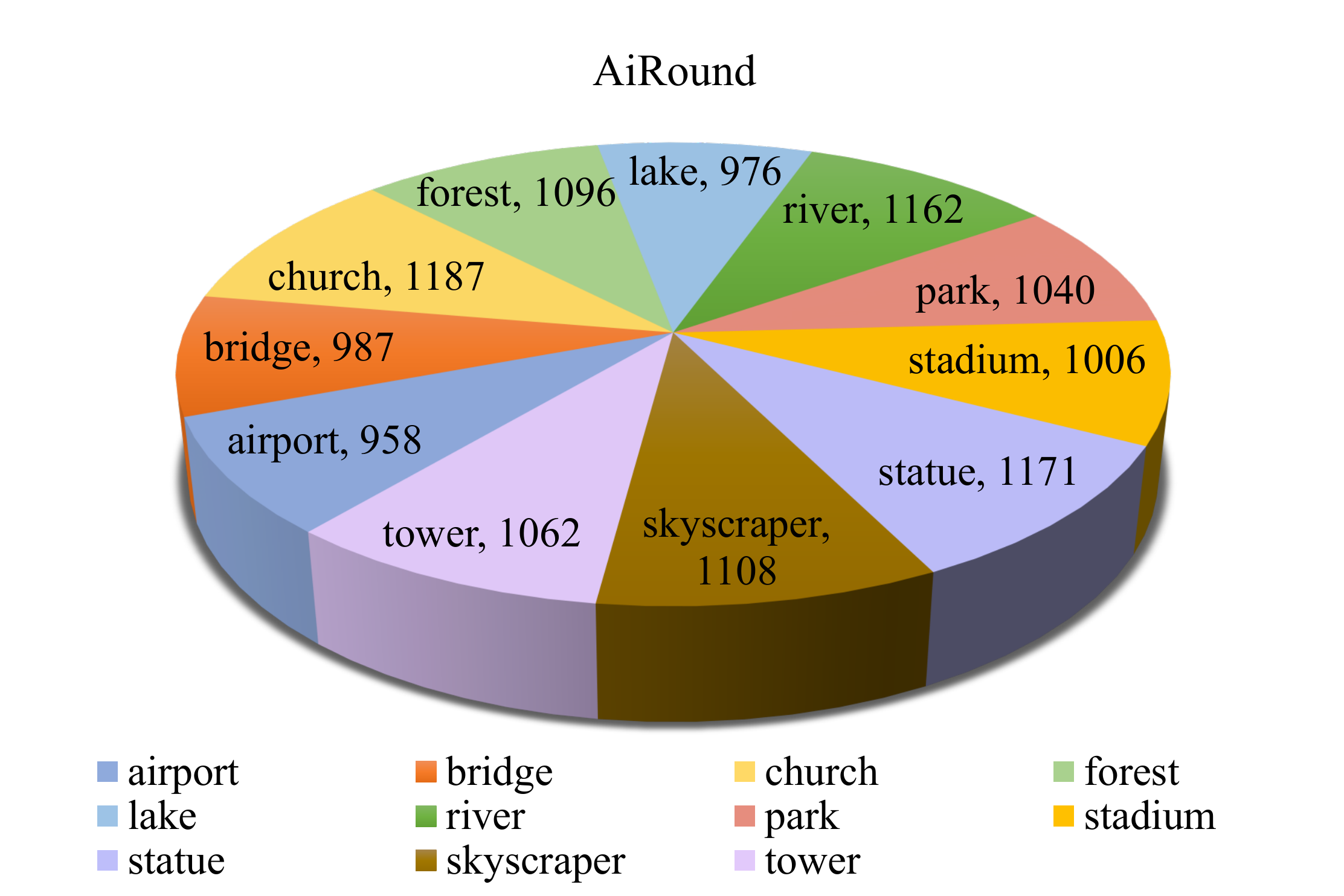}
\caption{Class distribution of the AiRound dataset.}
\label{fig:AiRound}
\end{figure}

\begin{figure}[ht]
\centering
\includegraphics[width=9cm]{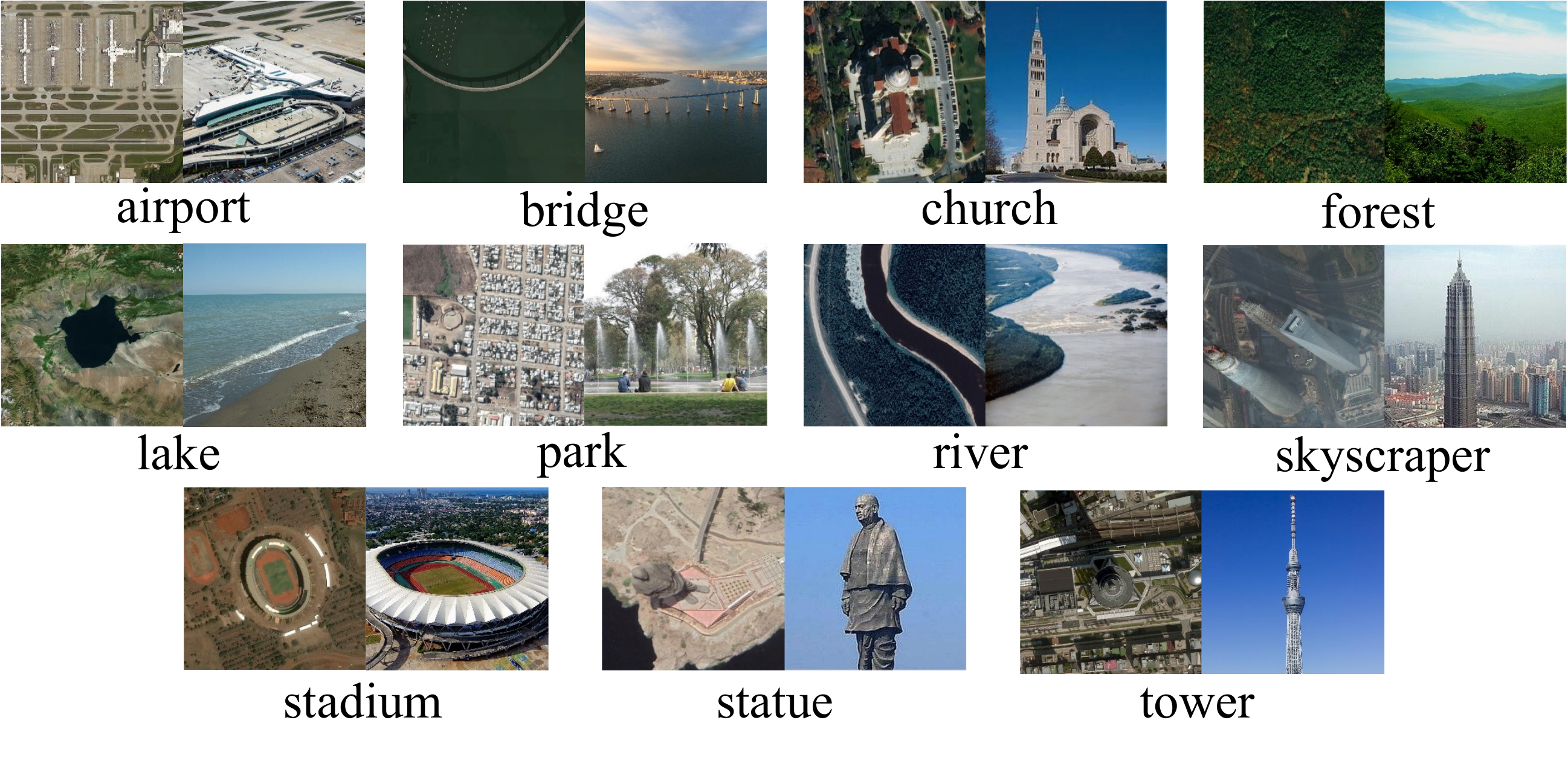}
\caption{Samples of the AiRound dataset.}
\label{fig:AiRound-Examples}
\end{figure}

For multi-class classification tasks, the cross-entropy loss function (CE Loss) is most commonly used. For a particular sample, its CE Loss can be calculated by

\begin{equation}
 L_{ce}= -\sum_{k=1}^K y_{k} \log p_{k},
\end{equation}
where $p_{k}$ is the predicted probability of the $k$-th class, $y_{k}$ is its class label and $y_{k}=1$ for positive classes and $y_{k}=0$ for negative classes. In the proposed model, the softmax operator is replaced with a non-negative activation function. Therefore, the CE Loss cannot be directly used. Since the evidence of this sample $\bm{e}$ obtained from the FEM can be mapped to the concentration parameters $\bm{\alpha}$ of the Dirichlet distribution $D(\mathbf{p}|\bm{\alpha})$ by using Eq.~\ref{eq:alpha}, the Bayes risk of cross-entropy loss $L_{cer}(\bm{\alpha})$ can be calculated as

\begin{equation}
\begin{aligned}
 L_{cer}(\bm{\alpha}) 
 & =\int L_{ce} D(\mathbf{p}|\bm{\alpha})d\mathbf{p}\\
 & =\int (-\sum_{k=1}^K y_{k}\log p_{k}) D(\mathbf{p}|\bm{\alpha})d\mathbf{p}.
\end{aligned}
\label{eq:ce-risk}
\end{equation}
Since $p_{k}$is a $D(\mathbf{p}|\bm{\alpha})$ random variable, the functions $\log p_{k}$ are the sufficient statistics of the Dirichlet distribution. Thus, the exponential family differential identities can be used to get an analytic expression for the expectation of $\log p_{k}$~\cite{lin2016dirichlet}:

\begin{figure}[ht]
\centering
\includegraphics[width=9cm]{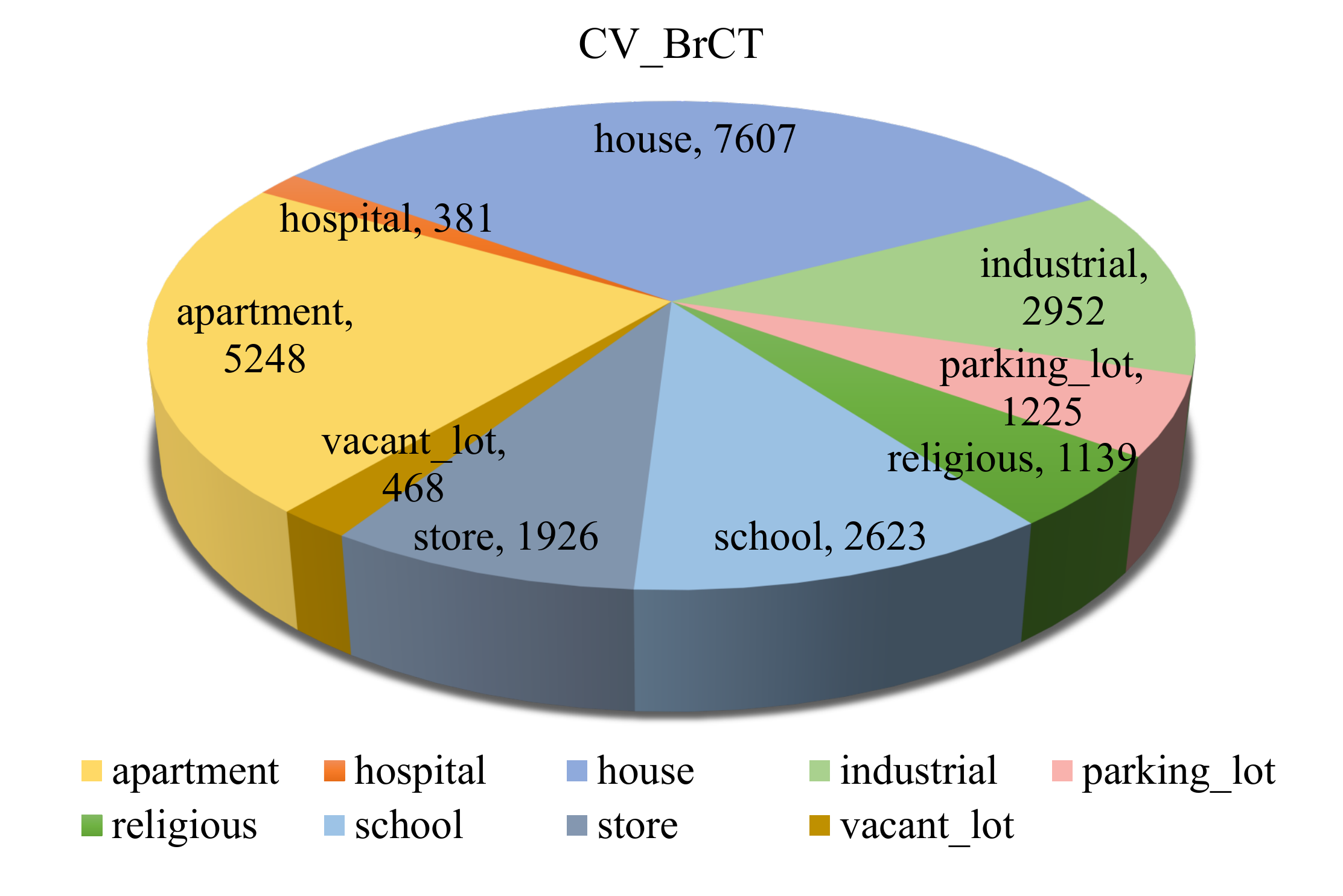}
\caption{Class distribution of the CV-BrCT dataset.}
\label{fig:CV-BrCT}
\end{figure}
 
\begin{figure}[ht]
\centering
\includegraphics[width=8cm]{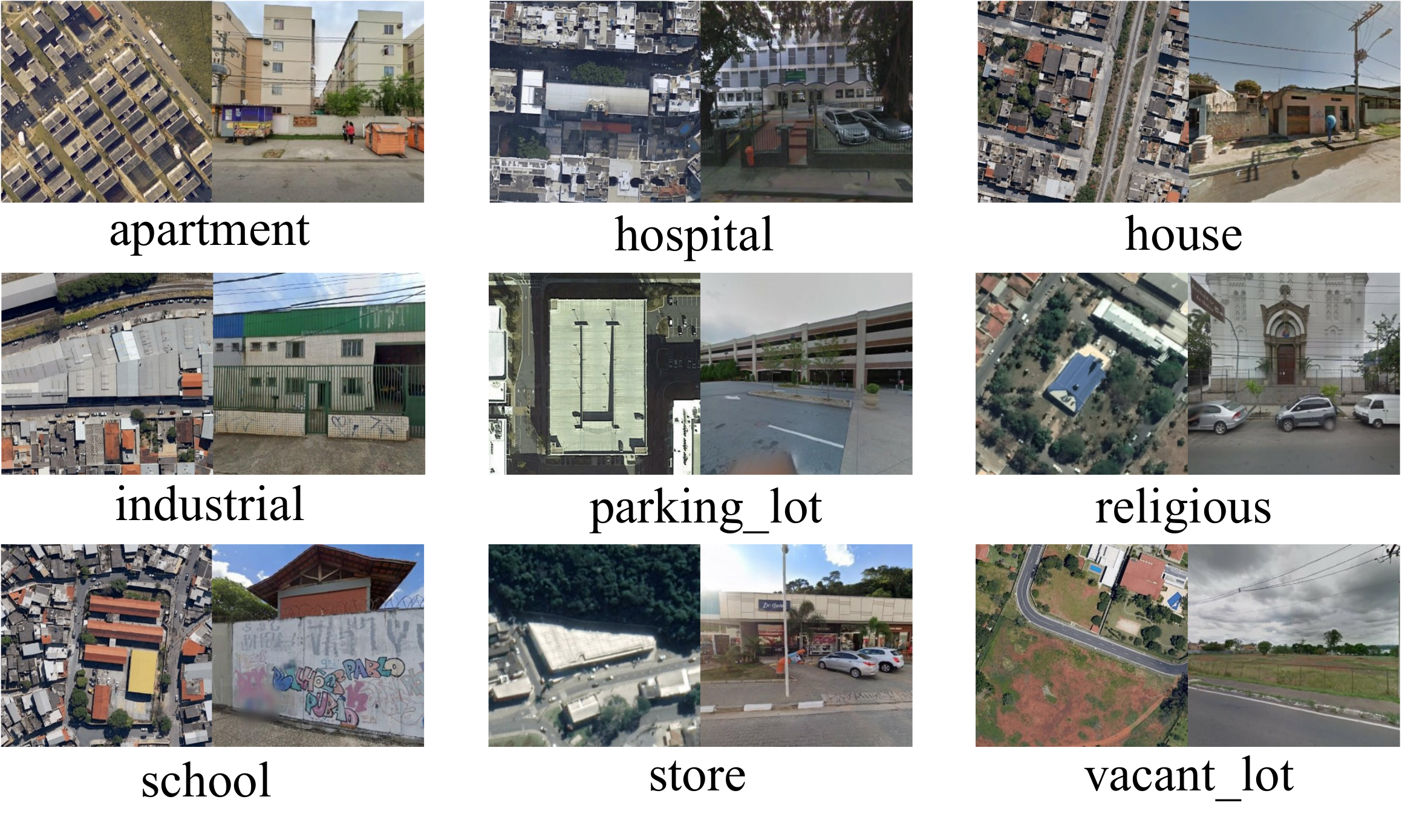}
\caption{Samples of the CV-BrCT dataset.}
\label{fig:CV-BrCT-Examples}
\end{figure}

\begin{equation}
\begin{aligned}
 \mathbb{E}_{D(\mathbf{p}|\bm{\alpha})}(\log p_{k}) 
 & = \int (\log p_{k}) D(\mathbf{p}|\bm{\alpha})d\mathbf{p}\\ 
 & = \psi(\alpha_{k})-\psi(\alpha_{0}).
\end{aligned}
\label{eq:E-logp}
\end{equation}
In this regard, Eq.~\ref{eq:ce-risk} can be expanded further as

\begin{equation}
\begin{aligned}
 L_{cer}(\bm{\alpha}) 
 & = \int (-\sum_{k=1}^K y_{k}\log p_{k}) D(\mathbf{p}|\bm{\alpha})d\mathbf{p}\\
 & = -\sum_{k=1}^K y_{k}\int (\log p_{k}) D(\mathbf{p}|\bm{\alpha})d\mathbf{p}\\
 & = -\sum_{k=1}^K y_{k} \left[ \psi(\alpha_{k})-\psi(\alpha_{0}) \right]\\
 & = \sum_{k=1}^K y_{k} \left[ \psi(\alpha_{0})-\psi(\alpha_{k}) \right].
\end{aligned}
\label{eq:ce-risk-fine}
\end{equation}
Given that the Eq.~\ref{eq:ce-risk-fine} can only penalize the positive class, $L_{cer}$ can be written as $L_{pc}$. We finally have $L_{pc}(\bm{\alpha}_{i})$ in Eq.~\ref{eq:pc-loss} for the case of all samples.

In order to ensure that both views can provide reasonable opinions for scene classification and thus improve the overall opinion after fusion, the final multi-view global Reciprocal Loss is used:
\begin{equation}
L_{global}=L^1+L^2+L_{fused},
\label{eq:loss-global}
\end{equation}
where $L^1$, $L^2$ are the Reciprocal Loss (Eq.~\ref{eq:overall-loss}) for the first view and second view respectively, and $L_{fused}$ is obtained using Eq.~\ref{eq:overall-loss} on the fused parameters by Eq.~\ref{eq:ck-fusion}, Eq.~\ref{eq:u-fusion}, Eq.~\ref{eq:ek-update} and Eq.~\ref{eq:alpha}.

\begin{figure*}[ht]
\centering
\includegraphics[width=14cm]{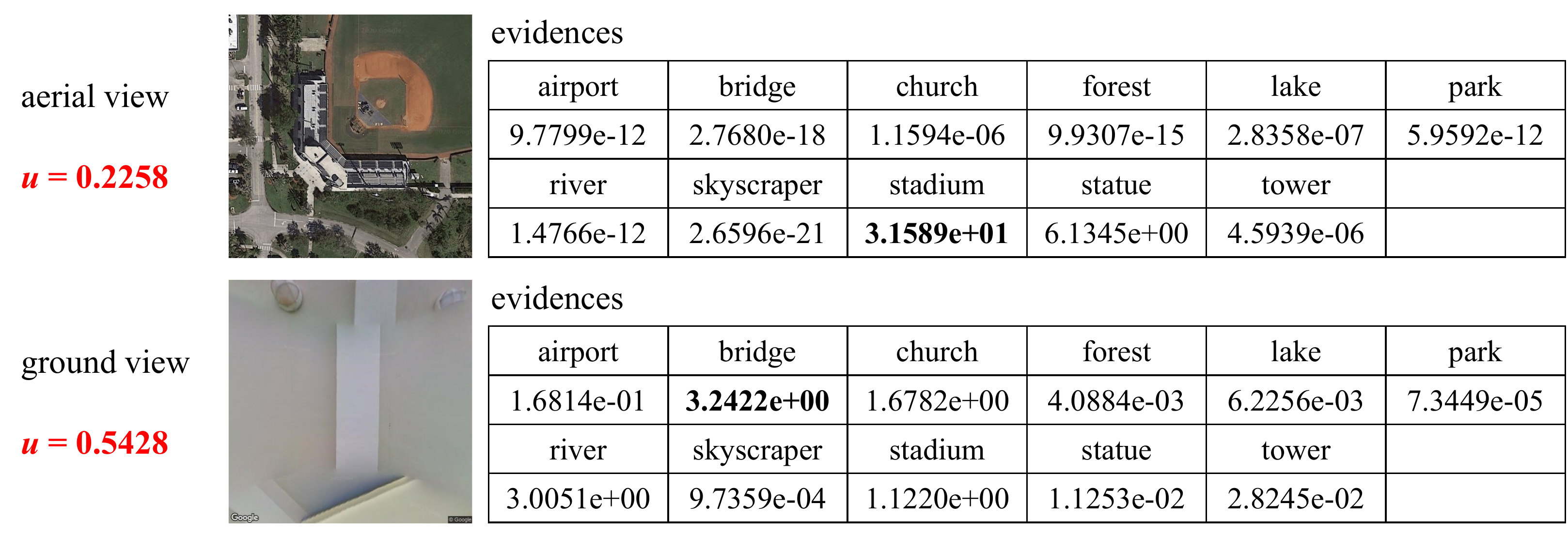}
\caption{The uncertainties and evidences of sample No.9360855 of class “stadium” in AiRound computed by the TEB-SUQ. The evidence values in bold correspond to the predicted results.}
\label{fig:Uncertainties+Evidences}
\end{figure*}

\section{Experiments}
\label{sec: Experiments}
\subsection{Datasets Description and Experimental Setup}
\label{subsec: Experimental Setup}
In this paper, performance of the proposed approach has been verified through experiments on two public aerial-ground dual-view images datasets~\cite{machado2020airound}. 

The AiRound dataset is a collection of landmarks from all over the world. It consists of images from three different views: the Sentinel-2 images, the high-resolution RGB aerial images, and the ground images. Sentinel-2 images have a size of 224×224 pixels. Aerial images have a size of 500×500 pixels. Ground images are obtained from two different ways, namley, Google Places’ database and Google Images. Thus they have different sizes. Their labels are obtained from the publicly available data of the OpenStreetMap. As shown in Fig.~\ref{fig:AiRound}, there are 11 different land use classes in AiRound with a total of 11,753 groups of images. The aerial and ground images are used in our experiments. Fig.~\ref{fig:AiRound-Examples} shows several samples.

The CV-BrCT dataset contains 23,569 pairs of images in 9 classes: apartment, hospital, house, industrial, parking\underline{ }lot, religious, school, store, and vacant\underline{ }lot. Each pairs are composed of an aerial view image and a ground view image, both of which are 500×500 RGB images. The class distribution and several samples are shown in Fig.~\ref{fig:CV-BrCT} and Fig.~\ref{fig:CV-BrCT-Examples}, respectively.

For both datasets, we randomly selected 80\% of the samples from each class as the training/validation set and the remaining 20\% as the test set to form one data split. All the test results except for Section~\ref{subsub:Ablation-Uncertainty-Estimation} are the mean results of 10 splits. The training/validation set was then randomly divided into training set and validation set according to the ratio of 9:1.

All models were simulated by PyTorch on a computer with a GTX 1080Ti graphics card. The details during training are as follows. Batch size: 128; the number of epochs: 200 for feature extraction and credible fusion, respectively, and the all best models on validation data are saved, learning rate schedules: Cosine decay with the initial value of 0.01, optimizer: SGD for feature extraction and Adam for credible fusion, weight decay: 0.1, and momentum of SGD: 0.9. All models of feature extraction are trained by fine-tuning the officially published pre-trained ones.

To quantitatively evaluate the performance of each model, we used the classification accuracy (Acc) and F1-score (F1) as metrics.

\subsection{Ablation Study}
\label{subsec:Ablation-Study}
\subsubsection{Validation for The Effectiveness of Uncertainty Estimation}
\label{subsub:Ablation-Uncertainty-Estimation}

Fig.~\ref{fig:Uncertainties+Evidences} shows one study case of the class “stadium” in AiRound. By the TEB-SUQ, the uncertainties and evidences of two images in different views are obtained. The evidence values bigger than 1.000 of the aerial iamge are 31.589 (stadium) and 6.135 (statue), shownig a good concentration. Accordingly, its uncertainty is only about 0.226. By contrast, the evidence values bigger than 1.000 of the ground iamge are 3.242 (bridge), 3.005 (river), 1.678 (church) and 1.122 (stadium), whose distribution is more dispersed. Accordingly, its uncertainty is about 0.543, suggesting that the model is less than half as confident about its predictions. As can be seen from the images, the above conclusions are intuitive.

\begin{figure}[ht]
\centering
\includegraphics[width=8cm]{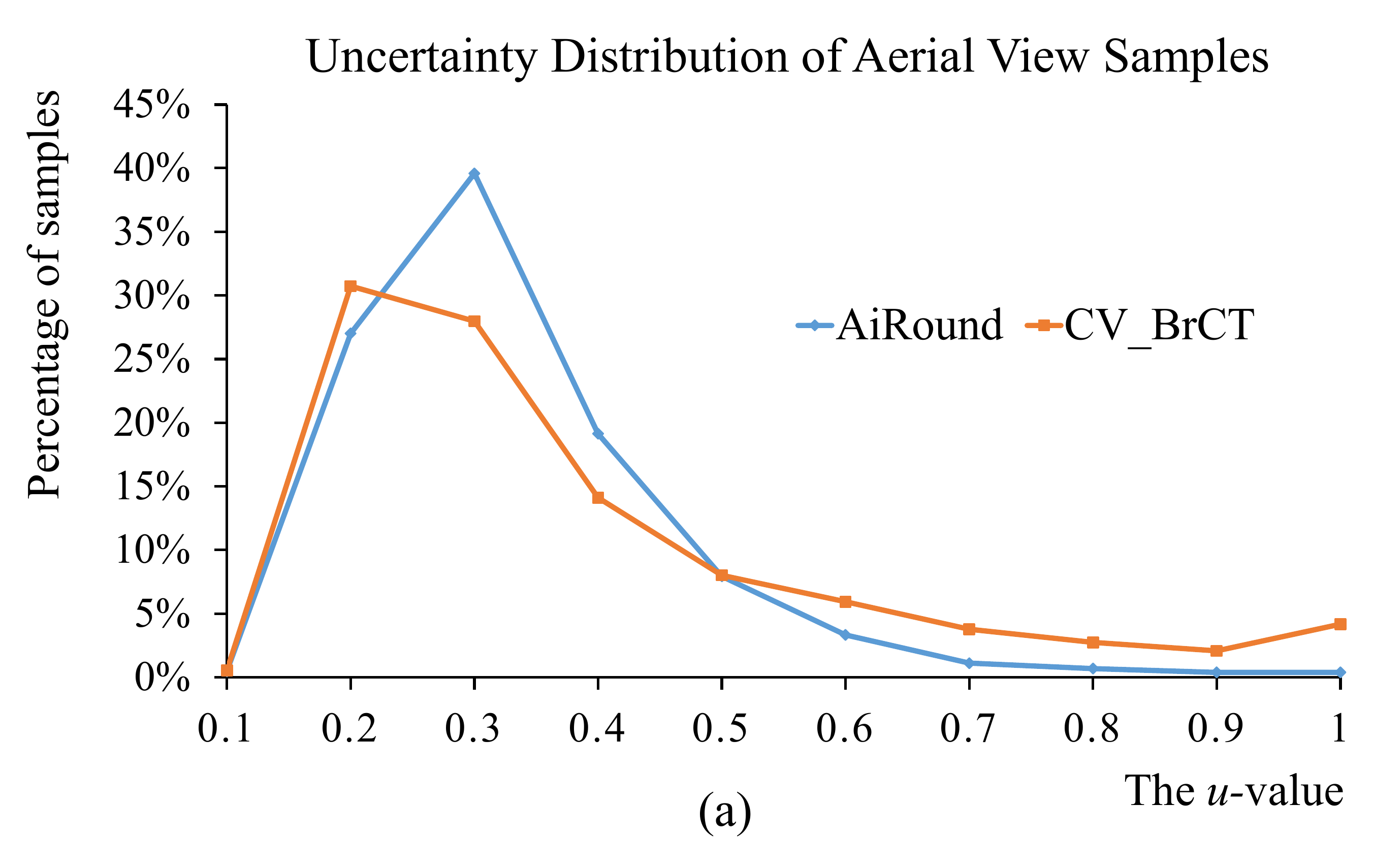}
\includegraphics[width=8cm]{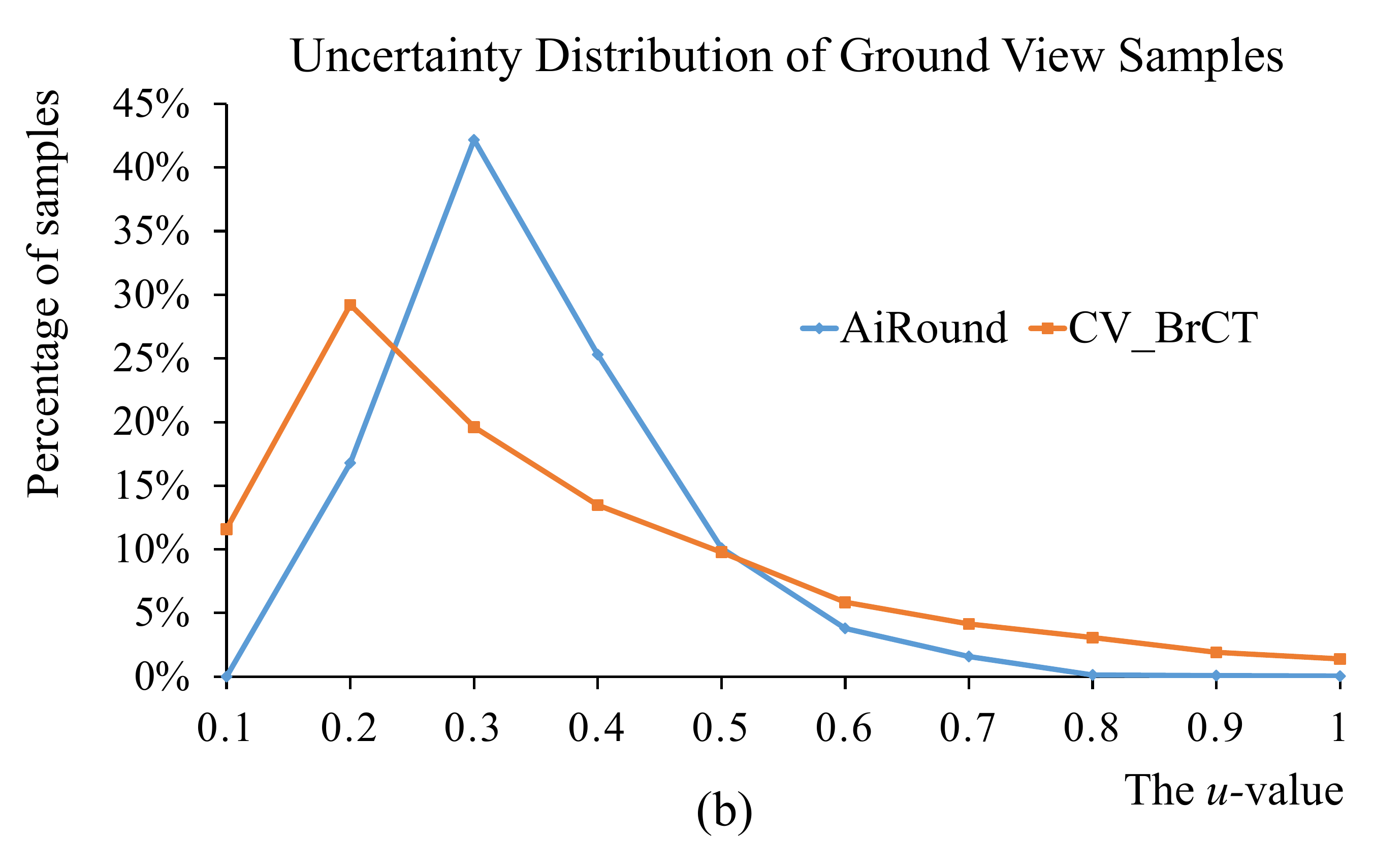}
\caption{The uncertainty distributions of the (a) aerial and (b) ground view samples in the test sets of AiRound and CV BrCT.}
\label{fig:Uncertainty-Distributions}
\end{figure}

More statistically, Fig.~\ref{fig:Uncertainty-Distributions} shows the uncertainty distributions of each view samples in the test sets of the AiRound and CV-BrCT. The figures clearly show that the samples of AiRound are distributed more densely in parts with lower uncertainty and have higher peak values. This advantage is even more visible in the ground view (Fig.~\ref{fig:Uncertainty-Distributions}(b)). In other words, after the calculation by the TEB-SUQ, the quality of AiRound is higher than that of CV-BrCT, especially in the ground view. This conclusion is supported by the following facts. Firstly, CV-BrCT has more than twice the number of samples as AiRound. Secondly, unlike the average distribution of AiRound, the class distribution of CV-BrCT is a typical long-tail distribution (Fig.~\ref{fig:CV-BrCT}). It is well understood that increasing the number of samples and unbalanced category distribution will result in a decrease in data quality. Last but not least, the methodologies used to collect the ground view images for the two datasets differ~\cite{machado2020airound}. The ground view images of AiRound are largely derived from the Google Places’ database, which is a well known high-quality dataset. The ground view images of CV-BrCT, on the other hand, are all obtained by the Google Images search engine, which cannot guarantee the image quality.

\begin{figure}[ht]
\centering
\includegraphics[width=8cm]{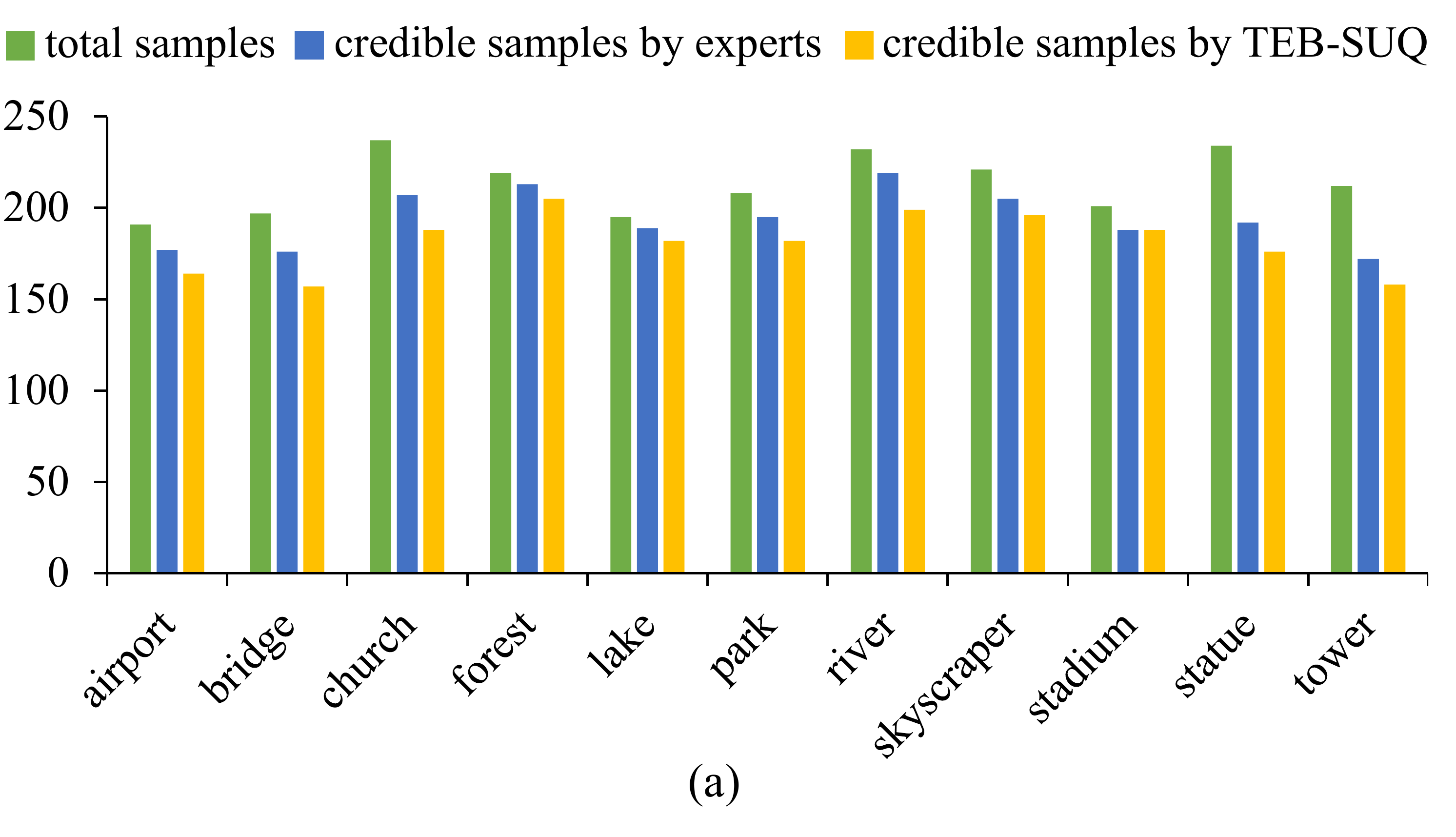}
\includegraphics[width=8cm]{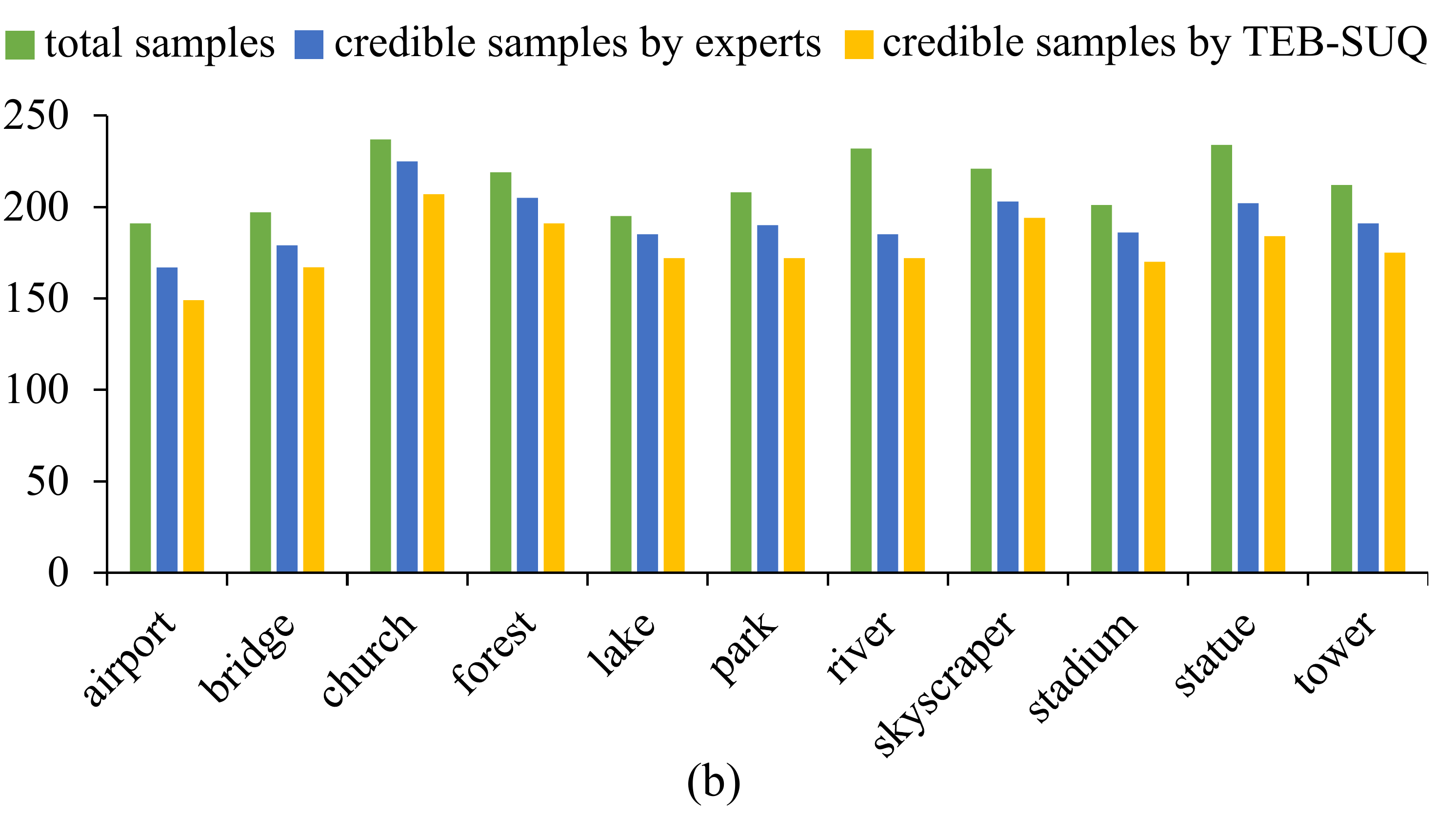}
\includegraphics[width=8cm]{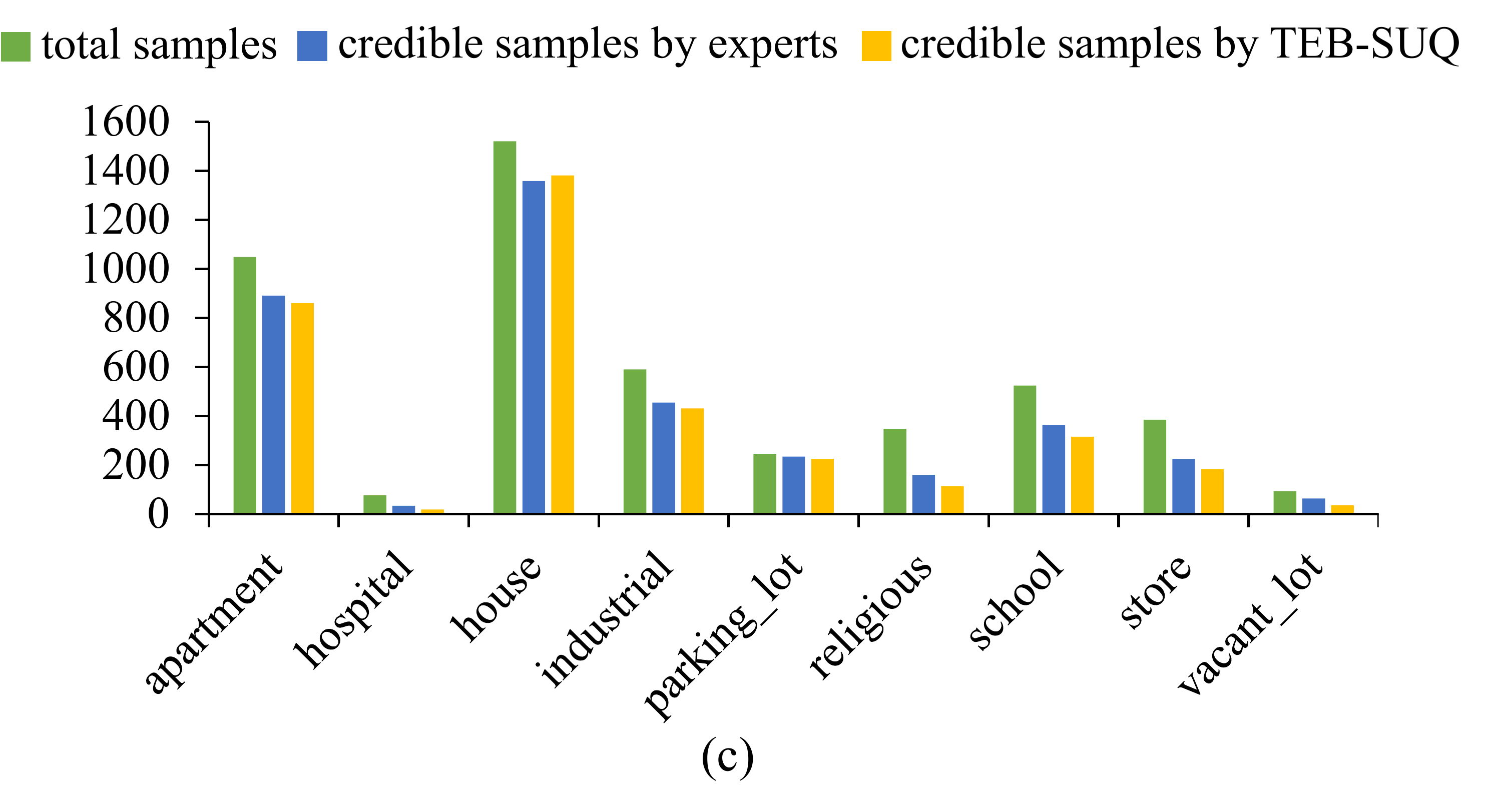}
\includegraphics[width=8cm]{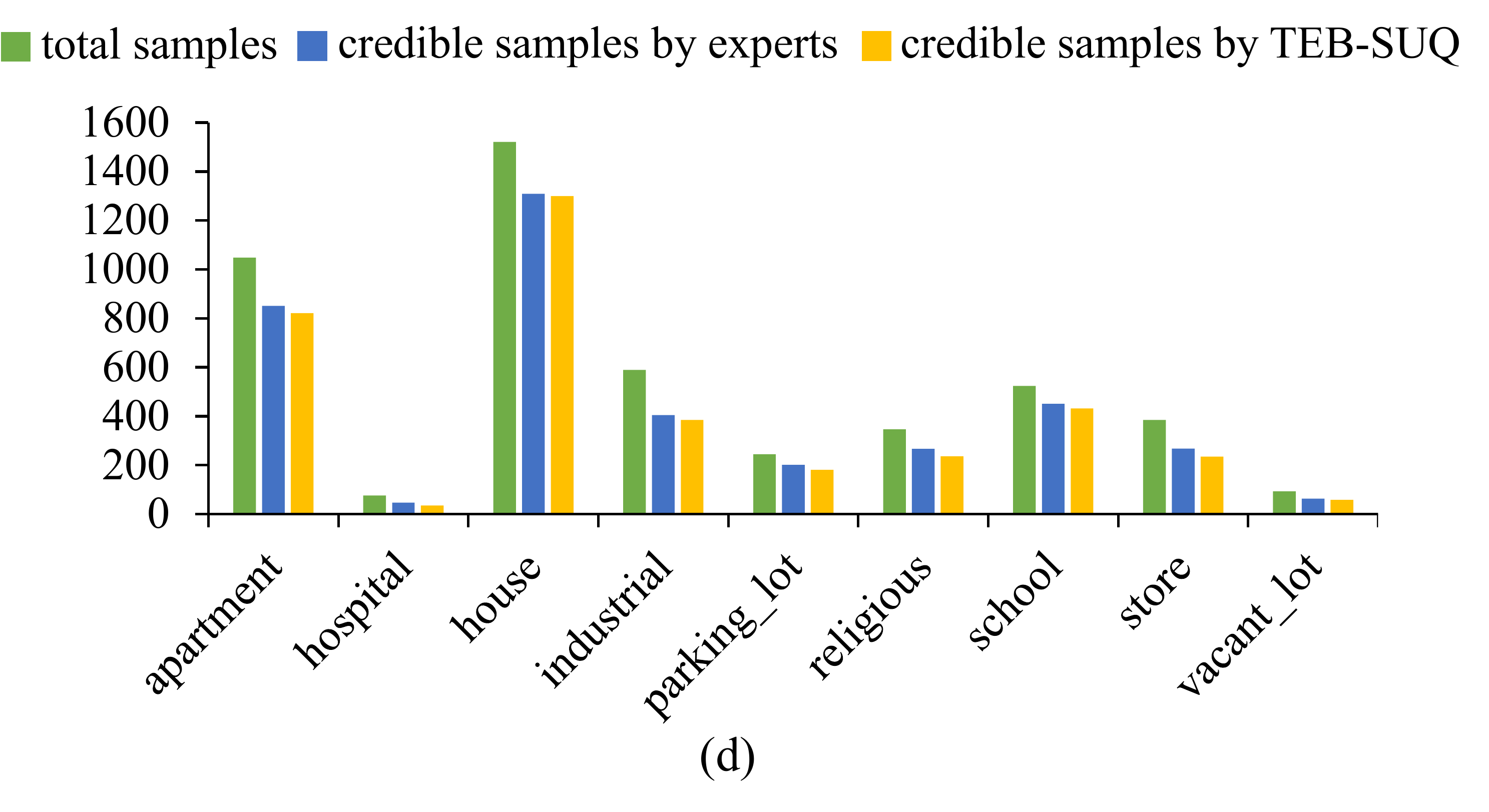}
\caption{The number of credible samples in each class generated by experts’ votes and the TEB-SUQ using $u\leq 0.4$ in the (a) aerial and (b) ground view of AiRound; (c) aerial and (d) ground view of CV-BrCT.}
\label{fig:Uncertainty-Validation}
\end{figure}

In order to further validate the effectiveness of the TEB-SUQ in measuring sample uncertainty, a subjective evaluation of sample credibility was performed. All test samples from the two datasets (2,347 pairs of AiRound and 4,830 pairs of CV-BrCT) were evaluated by 9 urban planning experts from the Qingdao Research Institute of Urban and Rural Construction based on whether their image content matched the label. A vote of 9 experts determines the final conclusion (credible or uncertain) of each test sample. For the TEB-SUQ, an appropriate threshold $u_{th}$ is set to distinguish between credible and uncertain samples. The number of credible samples calculated using $u\leq 0.4$ ($n_{k}$ for the $k$-th class) is compared to the one generated by experts’ votes ($m_{k}$ for the $k$-th class) in each class of each view of the two datasets in Fig.~\ref{fig:Uncertainty-Validation}. TABLE~\ref{tab:average-error} shows the average error $\bar{\sigma}$ of the TEB-SUQ relative to subjective evaluation of each view using
\begin{equation}
\bar{\sigma} = \frac{1}{K} \sum_{k=1}^K \frac{\lvert n_{k}-m_{k} \rvert}{m_{k}},
\label{eq:average-error}
\end{equation}
where $K$ is the total classes.

\begin{table}[h]
	\centering
	\caption{The average error of the TEB-SUQ relative to subjective evaluation.}
	\setlength{\tabcolsep}{7.5pt}
	\label{tab:average-error}
	\begin{tabular}{ccccc}
		\Xhline{1pt}
        \noalign{\smallskip}
		\textbf{Views}* & \textbf{A-}\textit{AiRound} & \textbf{G-}\textit{AiRound} & \textbf{A-}\textit{CV-BrCT} & \textbf{G-}\textit{CV-BrCT} \\
        \noalign{\smallskip}
		\Xhline{0.5pt}
		\noalign{\smallskip}
        $\bar{\sigma}$& 0.065 & 0.078 & 0.183 & 0.090 \\
		\noalign{\smallskip}
		\Xhline{1pt}
	\end{tabular}
        \begin{tablenotes}
        \footnotesize
        \item*\textbf{A-} for the aerail view and \textbf{G-} for the ground view.
        \end{tablenotes}
\end{table}

\begin{figure*}[ht]
\centering
\includegraphics[width=14cm]{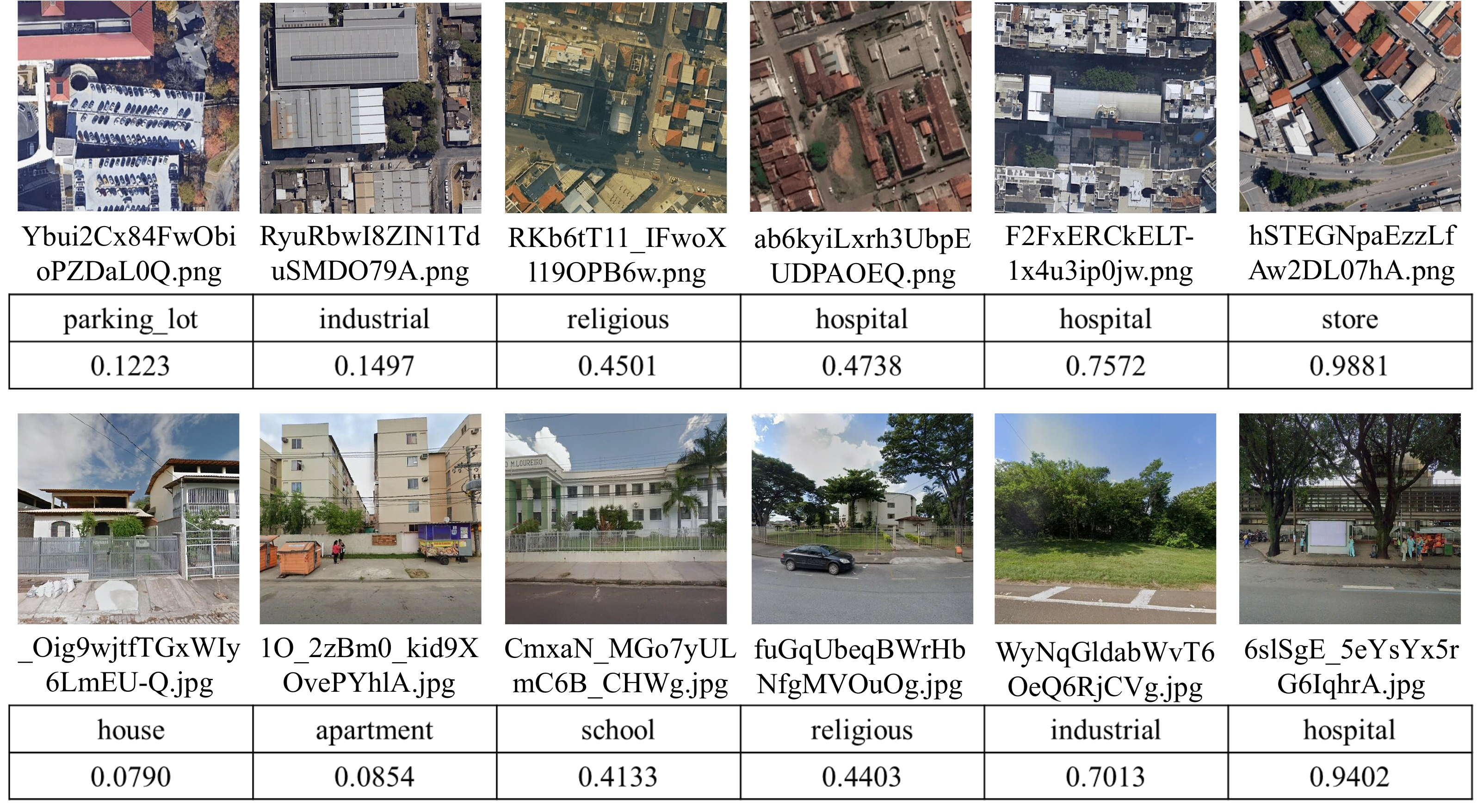}
\caption{The uncertainty of samples in CV-BrCT computed by the TEB-SUQ.}
\label{fig:More-Cases}
\end{figure*}

As can be observed from TABLE~\ref{tab:average-error}, except for the aerial view of CV-BrCT, the relative error between the predicted number of credible samples and the subjective evaluation in other views is less than 0.1, proving that the proposed sample uncertainty estimation approach is effective. More cases of the uncertainty of samples in the CV-BrCT datasets are shown in Fig.~\ref{fig:More-Cases}, whose classes are random selected.

\begin{figure*}[ht]
\centering
\includegraphics[width=16cm]{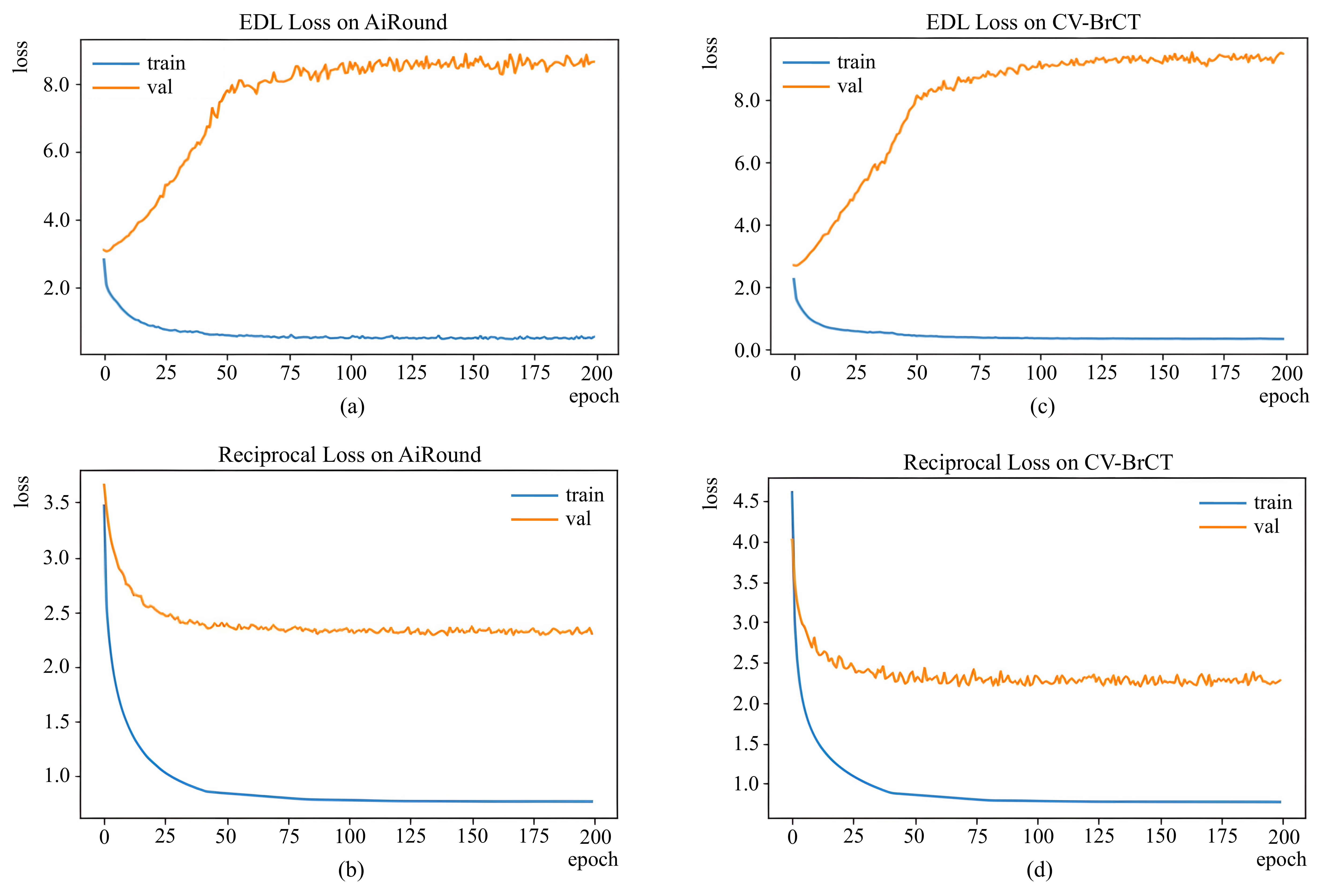}
\caption{The training and validation loss using (a) EDL Loss and (b) Reciprocal Loss on AiRound; (c) EDL Loss and (d) Reciprocal Loss on CV-BrCT.}
\label{fig:Loss-Validation}
\end{figure*}

\subsubsection{Validation for The Effectiveness of The Reciprocal Loss}
\label{subsub:Ablation-Loss}
The widely used loss function for evidential deep learning (EDL Loss for short) consists of a term of CE-like Loss and a term of Kullback-Leibler (KL) divergence of two Dirichlet distribution with a extra annealing coefficient~\cite{sensoy2018evidential}. It is formally too complex and difficult to train, in contrast to the proposed Reciprocal Loss (Eq.~\ref{eq:overall-loss} to Eq.~\ref{eq:nc-loss}). Fig.~\ref{fig:Loss-Validation} shows the training and validation loss using EDL Loss and Reciprocal Loss (Eq.~\ref{eq:loss-global}) with the same setting. Significant overfitting occurred during training using EDL Loss. It has been significantly improved since switching to Reciprocal Loss. Scene classification performances of EDL Loss and the proposed Reciprocal Loss are shown in TABLE~\ref{tab:Loss-Validation}. All results are obtained using VGG-11~\cite{simonyan2014very} as the backbone with the same training setting and fusion strategy. The annealing coefficient in EDL is set to 50. It is clear that the optimization of the training processes based on Reciprocal Loss resulted in improved performances in the test set. 


\begin{table}[h]
	\centering
	\caption{Performances (\%) {$\rm{\pm{~STD}}$} of different loss function.}
	\setlength{\tabcolsep}{2.5pt}
	\label{tab:Loss-Validation}
	\begin{tabular}{ccccccc}
 
		\Xhline{1pt}
        \noalign{\smallskip}
        
		\textbf{Loss} &~~& \multicolumn{2}{c}{\textbf{AiRound}} &~~& \multicolumn{2}{c}{\textbf{CV-BrCT}} \\
  
		\Xcline{3-4}{0.2pt} \Xcline{6-7}{0.2pt}
		\noalign{\smallskip}
  
		\textbf{Function} &~~& \textbf{Acc} & \textbf{F1} &~~& \textbf{Acc} & \textbf{F1} \\
                               
        \Xhline{0.5pt}
		\noalign{\smallskip}
  
		EDL Loss &~~& 90.23$\pm{0.32}$ & 90.64$\pm{0.29}$ &~~& 86.98$\pm{0.17}$ & 81.56$\pm{0.34}$\\
        \noalign{\smallskip}
        Reciprocal Loss &~~& \textbf{92.16$\pm$0.31} & \textbf{92.49$\pm$0.25} &~~& \textbf{88.21$\pm$0.26} & \textbf{83.57$\pm$0.29}\\
  
		\noalign{\smallskip}
		\Xhline{1pt}
	\end{tabular}
\end{table}

\subsubsection{Validation for The Effectiveness of The Evidential Fusion Strategy}
\label{subsub:Ablation-Fusion}

As described in Section~\ref{subsec: Overview}, the proposed approach is backbone independent, allowing it to be flexibly matched with different deep feature extraction networks. In this experiment, three of the most common backbones are used to compare the performance on scene classification task before and after using different fusion strategies. In the columns of single views in TABLE~\ref{tab:Fusion-Validation}, the “-s” represents the traditional deep learning (that is, softmax deep learning) approach where the softmax layer and CE Loss are used during the training phase, and the class corresponding to the maximum probability calculated by the softmax operator is used as the prediction result during the test phase. And “-e” denotes the evidential deep learning approach proposed in Section~\ref{subsec: Uncertainty Estimation} and Section~\ref{subsec: Loss Function} where the softmax layer is replaced by the softplus layer and the proposed Reciprocal Loss are used for training. The class corresponding to the maximum evidence value is used as the prediction result during test. In the columns of fusion strategies, two common decision level fusion strategies (sum and product)~\cite{machado2020airound} are used as baseline approaches to compare the proposed evidential fusion. The best results are highlighted in bold. 

The following observations can be made from TABLE~\ref{tab:Fusion-Validation}. First, all of the fusion results are superior to any single view result. This demonstrates that information from multiple perspectives can significantly improve classification accuracy. Second, the performance of the two single views obtained through evidential deep learning is marginally worse than that of the corresponding single view obtained through softmax deep learning. This is due to the fact that when the uncertainty of some samples is high, evidential deep learning focuses more on estimating the uncertainty values as accurately as possible, which may result in a loss of classification accuracy for these samples. However, in terms of sample quality, the category labels of these high-uncertainty samples lack actual semantics. It also doesn't matter whether their predictions are correct or incorrect. It makes more sense to quantify their uncertainty. Last but not least, the evidential fusion strategy proposed in this paper outperforms the other two baseline fusion approaches, which demonstrates the efficacy of evidential fusion in the task of multi-view remote sensing scene classification.

\begin{table*}[ht]
	\centering
	\caption{Classification accuracy (\%) {$\rm{\pm{~STD}}$} before and after decision-level fusion.}
	\setlength{\tabcolsep}{3.5pt}
	\label{tab:Fusion-Validation}
	\begin{tabular}{ccrcccccccccc}
 
		\Xhline{1pt}
        \noalign{\smallskip}
        
		\multirow{2}{*}{\textbf{Dataset}} & \multirow{2}{*}{~~} & \multirow{2}{*}{\textbf{Backbone}} & \multirow{2}{*}{~~} & \multicolumn{2}{c}{\textbf{Single Views (softmax)}} & ~~ & \multicolumn{2}{c}{\textbf{Single Views (evidential)}} & ~~ & \multicolumn{3}{c}{\textbf{Decision-Level Fusion Strategies}} \\
  
		\Xcline{5-6}{0.2pt} \Xcline{8-9}{0.2pt} \Xcline{11-13}{0.2pt}
		\noalign{\smallskip}

        & & & & \textbf{Aerial-s} & \textbf{Ground-s} & ~~ & \textbf{Aerial-e} & \textbf{Ground-e} & ~~ & \textbf{Sum}\cite{machado2020airound} & \textbf{Product}\cite{machado2020airound} & \textbf{Evidential (proposed)}\\
                               
        \Xhline{0.5pt}
		\noalign{\smallskip}
  
		\multirow{3}{*}{Airound} & \multirow{3}{*}{~~} & AlexNet\cite{krizhevsky2017imagenet} & ~~ & 76.96$\pm{0.52}$ & 71.35$\pm{0.24}$ & ~~ & 76.04$\pm{0.46}$ & 70.96$\pm{0.21}$ & ~~ & 84.02$\pm{0.47}$ & 86.74$\pm{0.25}$ & \textbf{88.12$\pm$0.23} \\
        \noalign{\smallskip}
        & & VGG-11\cite{simonyan2014very} & ~~ & 82.75$\pm{0.61}$ & 77.10$\pm{0.28}$ & ~~ & 82.64$\pm{0.49}$ & 76.99$\pm{0.17}$ & ~~ & 87.75$\pm{0.38}$ & 90.41$\pm{0.27}$ & \textbf{92.16$\pm$0.31} \\
        \noalign{\smallskip}
        & & ResNet-18\cite{he2016deep} & ~~ & 80.93$\pm{0.49}$ & 76.68$\pm{0.19}$ & ~~ & 80.83$\pm{0.52}$ & 76.36$\pm{0.22}$ & ~~ & 88.02$\pm{0.25}$ & 89.56$\pm{0.24}$ & \textbf{91.02$\pm$0.35} \\

        \noalign{\smallskip}
        \Xhline{0.5pt}
		\noalign{\smallskip}
  
		\multirow{3}{*}{CV-BrCT} & \multirow{3}{*}{~~} & AlexNet\cite{krizhevsky2017imagenet} & ~~ & 84.63$\pm{0.24}$ & 68.01$\pm{0.12}$ & ~~ & 84.37$\pm{0.10}$ & 66.36$\pm{0.25}$ & ~~ & 85.26$\pm{0.45}$ & 86.52$\pm{0.25}$ & \textbf{88.02$\pm$0.28} \\
        \noalign{\smallskip}
        & & VGG-11\cite{simonyan2014very} & ~~ & 87.11$\pm{0.42}$ & 71.43$\pm{0.22}$ & ~~ & 87.06$\pm{0.38}$ & 70.15$\pm{0.29}$ & ~~ & 86.70$\pm{0.58}$ & 87.21$\pm{0.22}$ & \textbf{88.21$\pm$0.26} \\
        \noalign{\smallskip}
        & & ResNet-18\cite{he2016deep} & ~~ & 86.74$\pm{0.38}$ & 70.96$\pm{0.25}$ & ~~ & 86.18$\pm{0.29}$ & 70.86$\pm{0.24}$ & ~~ & 85.59$\pm{0.62}$ & 86.83$\pm{0.18}$ & \textbf{87.95$\pm$0.19} \\
  
		\noalign{\smallskip}
		\Xhline{1pt}
	\end{tabular}
\end{table*}

\begin{table*}[ht]
	\centering
	\caption{Classification accuracy (\%) {$\rm{\pm{~STD}}$} using different fusion strategies.}
	\setlength{\tabcolsep}{1.9pt}
	\label{tab:Comparison-Experiment}
	\begin{tabular}{ccrccccccccccc}
 
		\Xhline{1pt}
        \noalign{\smallskip}
        
		\multirow{2}{*}{\textbf{Dataset}} & \multirow{2}{*}{~} & \multirow{2}{*}{\textbf{Backbone}} & \multirow{2}{*}{~} & \textbf{Data-Level} & ~ & \multicolumn{2}{c}{\textbf{Feature-Level}} & ~ & \multicolumn{5}{c}{\textbf{Decision-Level}} \\
  
		\Xcline{5-5}{0.2pt} \Xcline{7-8}{0.2pt} \Xcline{10-14}{0.2pt}
		\noalign{\smallskip}

        & & & & \textbf{Six-Ch.}\cite{vo2016localizing} & ~ & \textbf{Concat.}\cite{machado2020airound} & \textbf{CILM}\cite{geng2022multi} & ~ &  \textbf{Max.}\cite{machado2020airound} & \textbf{Min.}\cite{machado2020airound} & 
        \textbf{Sum}\cite{machado2020airound} & \textbf{Product}\cite{machado2020airound} & \textbf{Evidential (proposed)}\\
                               
        \Xhline{0.5pt}
		\noalign{\smallskip}
  
		\multirow{5}{*}{Airound} & \multirow{5}{*}{~} & AlexNet\cite{krizhevsky2017imagenet} & ~ & 70.19$\pm{0.23}$ & ~ & 82.52$\pm{0.32}$ &  83.49$\pm{0.17}$ & ~ & 84.86$\pm{0.36}$ &  85.52$\pm{0.23}$ & 84.02$\pm{0.47}$ & 86.74$\pm{0.25}$ & \textbf{88.12$\pm$0.23} \\
        \noalign{\smallskip}
        & & VGG-11\cite{simonyan2014very} & ~ & 72.34$\pm{0.21}$ & ~ & 84.69$\pm{0.41}$ &  85.72$\pm{0.19}$ & ~ & 88.17$\pm{0.34}$ &  89.56$\pm{0.25}$ &  87.75$\pm{0.38}$ & 90.41$\pm{0.27}$ & \textbf{92.16$\pm$0.31} \\
        \noalign{\smallskip}
        & & Inception\cite{szegedy2016rethinking} & ~ & 71.76$\pm{0.24}$ & ~ & 83.91$\pm{0.45}$ & 85.05$\pm{0.15}$ & ~ & 88.39$\pm{0.33}$ & 89.12$\pm{0.27}$ & 88.05$\pm{0.30}$ & 90.02$\pm{0.14}$ & \textbf{91.41$\pm$0.18} \\
         \noalign{\smallskip}
        & & ResNet-18\cite{he2016deep} & ~ & 71.29$\pm{0.26}$ & ~ & 83.56$\pm{0.39}$ & 84.72$\pm{0.21}$ & ~ & 88.21$\pm{0.38}$ & 88.42$\pm{0.22}$ & 88.02$\pm{0.25}$ & 89.56$\pm{0.24}$ & \textbf{91.02$\pm$0.35} \\
         \noalign{\smallskip}
        & & DenseNet\cite{huang2017densely} & ~ & 71.57$\pm{0.25}$ & ~ & 83.72$\pm{0.42}$ & 84.91$\pm{0.20}$ & ~ & 89.96$\pm{0.35}$ & 90.41$\pm{0.24}$ & 89.88$\pm{0.31}$ & 91.16$\pm{0.17}$ & \textbf{92.16$\pm$0.19} \\

        \noalign{\smallskip}
        \Xhline{0.5pt}
		\noalign{\smallskip}
  
		\multirow{5}{*}{CV-BrCT} & \multirow{5}{*}{~~} & AlexNet\cite{krizhevsky2017imagenet} & ~ & 71.92$\pm{0.26}$ & ~ & 81.86$\pm{0.42}$ &  83.10$\pm{0.20}$ & ~ & 85.52$\pm{0.24}$ &  86.02$\pm{0.21}$ & 85.26$\pm{0.45}$ & 86.52$\pm{0.25}$ & \textbf{88.02$\pm$0.28} \\
        \noalign{\smallskip}
        & & VGG-11\cite{simonyan2014very} & ~ & 73.46$\pm{0.24}$ & ~ & 83.25$\pm{0.39}$ &  84.32$\pm{0.19}$ & ~ & 86.74$\pm{0.39}$ & 86.95$\pm{0.27}$ & 86.70$\pm{0.58}$ & 87.21$\pm{0.22}$ & \textbf{88.21$\pm$0.26}\\
        \noalign{\smallskip}
        & & Inception\cite{szegedy2016rethinking} & ~ & 75.26$\pm{0.25}$ & ~ & 84.65$\pm{0.38}$ & 85.22$\pm{0.16}$ & ~ & 86.70$\pm{0.41}$ & 86.95$\pm{0.28}$ & 86.24$\pm{0.35}$ & 87.02$\pm{0.21}$ & \textbf{88.21$\pm$0.23} \\
         \noalign{\smallskip}
        & & ResNet-18\cite{he2016deep} & ~ & 73.25$\pm{0.28}$ & ~ & 83.28$\pm{0.41}$ & 84.31$\pm{0.17}$ & ~ & 85.84$\pm{0.43}$ & 86.24$\pm{0.25}$ & 85.59$\pm{0.62}$ & 86.83$\pm{0.18}$ & \textbf{87.95$\pm$0.19} \\
         \noalign{\smallskip}
        & & DenseNet\cite{huang2017densely} & ~ & 74.19$\pm{0.27}$ & ~ & 84.24$\pm{0.40}$ & 85.19$\pm{0.22}$ & ~ & 86.95$\pm{0.38}$ & 87.02$\pm{0.25}$ & 86.85$\pm{0.39}$ & 87.54$\pm{0.17}$ & \textbf{88.34$\pm$0.20} \\
  
		\noalign{\smallskip}
		\Xhline{1pt}
	\end{tabular}
\end{table*}

\begin{figure*}[h]
\centering
\includegraphics[width=16cm]{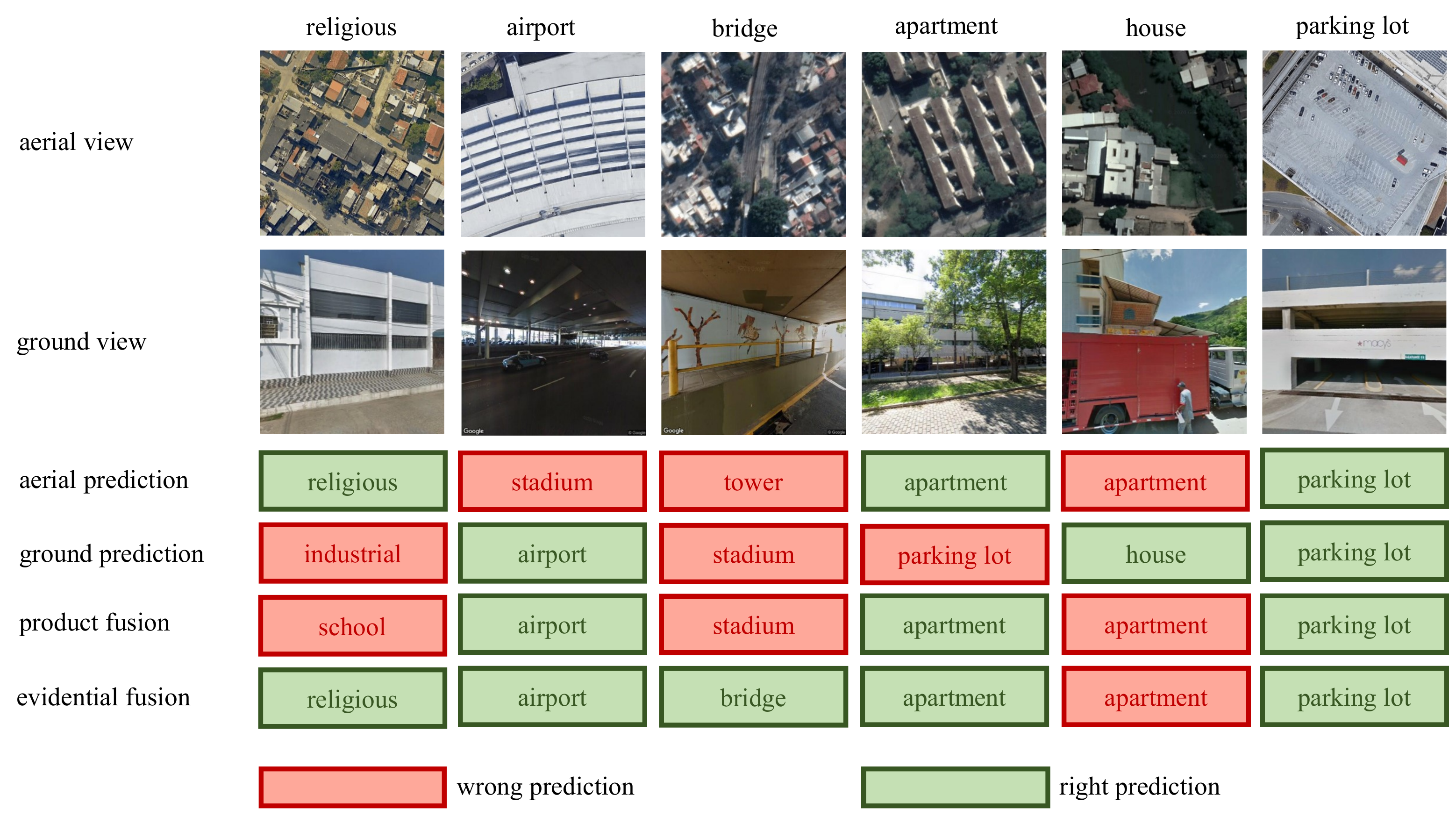}
\caption{Examples of predictions by single views, the product fusion and the proposed evidential fusion.}
\label{fig:Predictions-Cases}
\end{figure*}

\subsection{Comparison Experiment with Different Fusion Approaches at Data-Level, Feature-Level and Decision-Level}
\label{subsec: Comparison Experiment}

In Section~\ref{subsec:Ablation-Study}, the effectiveness of three innovative contributions in this paper (namely TEB-SUQ, Evidential Fusion, and Reciprocal Loss) was validated respectively. In this section, more multi-view fusion methods are compared with the proposed approach to assess its overall performance on the task of aerial-ground dual-view remote sensing scene classification. As mentioned in Section~\ref{subsec: Multi-view Data Fusion in Remote Sensing}, existing multi-view fusion methods can be roughly classified as data-level, feature-level, and decision-level. In this experiment, one data-level fusion method (six-channel~\cite{vo2016localizing}), two feature-level fusion methods (feature concatenation~\cite{machado2020airound} and CILM~\cite{geng2022multi}), and four decision-level fusion methods~\cite{machado2020airound} (maximum, minimum, sum and product) are chosen to compare with the proposed evidential fusion. These methods are briefly described below.
\begin{itemize}
\item 
\textit{Six-channel}~\cite{vo2016localizing}: This method concatenates the RGB channels of the paired aerial view and ground view images into a six-channel image as the input of a CNN. 

\item 
\textit{Feature concatenation}~\cite{machado2020airound}: This method uses a Siamese-like CNN to concatenate the intermediate feature vectors before the first convolution layer that doubles its amount of kernels. 

\item 
\textit{CILM}~\cite{geng2022multi}: The Loss function of contrast learning is combined with CE Loss in this method, allowing the features extracted by the two subnetworks to be fused without sharing any weight. 

\item 
\textit{Maximum}~\cite{machado2020airound}: Each view employs an independent DNN to obtain its prediction result, which consists of a class label and its probability. The final prediction is the class label corresponding to the maximum of the class probabilities predicted by each view. 

\item 
\textit{Minimum}~\cite{machado2020airound}: Each view employs an independent DNN to obtain its prediction result, which consists of a class label and its probability. The final prediction is the class label corresponding to the minimum of the class probabilities predicted by each view. 

\item 
\textit{Sum}~\cite{machado2020airound}: Each view employs an independent DNN to generate a vector containing probabilities for each class. The fused vector is the sum of single view vectors. The final prediction result is the class label corresponding to the largest element in the fused vector.

\item 
\textit{Product}~\cite{machado2020airound}: Each view employs an independent DNN to generate a vector containing probabilities for each class. An elementwise multiplication is performed between single view vectors to obtain the fused vector. The final prediction result is the class label corresponding to the largest element in the fused vector.
\end{itemize}

TABLE~\ref{tab:Comparison-Experiment} shows the performance of the fusion methods discussed above when different backbones are used. The following observations can be made. First, methods of the data-level fusion class performed the worst, while methods of the decision-level fusion class won across the board. This may be because the visual features of the aerial view differ so greatly from those of the ground view. The performance of the fusion improves as the features involved in it become more abstract. This also explains why CILM outperforms feature concatenation among the two feature-level fusion methods: CILM lacks a shared weight structure, making it more similar to decision-level fusion in form. Second, among the decision-level fusion methods, sum and maximum perform nearly identically, and both perform slightly worse than minimum. This result may seem counter-intuitive at first. In fact, it confirms the overconfidence issue caused by the softmax mentioned in Section~\ref{subsec: Uncertainty Estimation} (see Fig.~\ref{fig:over-confidence} and Fig.~\ref{fig:binary-classification}): an overestimated prediction is more likely to be incorrect. Last but not least, product and the proposed evidential fusion stand out among all the fusion methods, and the latter outperforms the former. In fact, Eq.~\ref{eq:ck-fusion} can be seen as an enhancement of the product method. The inclusion of sample uncertainty breaks down the equality of views in the fusion: views with lower $u$ values are given more weight adaptively. 

Finally, on all backbones of both datasets, the proposed evidential fusion approach outperforms the best decision-level fusion method by $1.26\% \pm 0.27$, the best feature-level fusion method by $4.96\% \pm 1.46$ and the data-level fusion method by $17.04\% \pm 2.68$. Examples of predictions are shown in Fig.~\ref{fig:Predictions-Cases}.

\section{Conclusion}
\label{sec:Conclusion}

Multi-view data (for example, aerial-ground dual-view images) are increasingly used in various remote sensing tasks such as scene classification. However, as the number of views grows, the problem of data quality in the original single view becomes more apparent, making the effect of fusion far less than expected. Deep learning models are easily influenced by data quality, especially when dealing with massive amounts of data. Feifei Li’s team proposed the concept of “trustworthy AI”~\cite{liang2022advances}, with a focus on data quality. This issue is also very noticeable in aerial-ground multi-view image data. The quality of aerial view images is relatively high. However, because of the large field of shooting scale, it can be difficult for a class label to accurately describe all of the scene categories contained in an image, resulting in a large number of samples that are diverse within the class and similar between the classes. On the other hand, there are many low-quality samples in the ground view, such as the target being too large, severe occlusion, indoor shooting. Existing fusion methods are frequently rendered ineffective in the face of these challenges. 

In this paper, uncertainty theory is introduced to try to quantify the credibility of samples. Specifically, the class scores output by DNN are mapped to an “opinion space” (see Fig.~\ref{fig:multi-classification} by a designed Dirichlet distribution to obtain an overall uncertainty value for the input sample. On this basis, a decision-level multi-view fusion strategy is proposed to assign higher weights to views with lower decision risk. In addition, a novel loss function, the Reciprocal Loss, is proposed. When compared to the loss function commonly used in evidential deep learning, Reciprocal Loss is more concise, easier to train, and achieves better test performance. Based on the three innovations mentioned above, the proposed envidential fusion approach achieves the best performance on the two classical datasets in the task of remote sensing scene classification on aerial-ground dual-view images.

Focusing on data quality in multi-view tasks is a new area of research, and much work remains to be done. First, there are few publicly available multi-view datasets for remote sensing tasks. Large-scale, instance-level aligned remote sensing multi-view data sets are urgently needed for public release for related research. Furthermore, effective objective evaluation of sample uncertainty estimation is lacking. Datasets with sample quality annotation have yet to appear in the field of remote sensing. Finally, more explicit representation methods of sample uncertainty need to be further explored.

\section*{Acknowledgment}
The authors would like to thank Ruxuan Bi, Zhiwei He and Mengshuo Fan, the experts in urban planning from the BIM Research Center, Qingdao Research Institute of Urban and Rural Construction for their professional guidance on the subjective evaluation of sample credibility. Thanks to those who participated in the subjective evaluation: Chunting Zhao, Xiayue Wang, Qi Liu, Mingdong Ding, Jinhong Guo and Zhengnan Fang. This work has been supported by the National Natural Science Foundation of China (Grant No.62171247).

\ifCLASSOPTIONcaptionsoff
  \newpage
\fi

\bibliographystyle{ieeetr} 
\bibliography{ref} 

\begin{thebibliography}{10}

\bibitem{roy2020metric}
S.~Roy, E.~Sangineto, B.~Demir, and N.~Sebe, ``Metric-learning-based deep
  hashing network for content-based retrieval of remote sensing images,'' {\em
  IEEE Geoscience and Remote Sensing Letters}, vol.~18, no.~2, pp.~226--230,
  2020.

\bibitem{wang2016three}
Y.~Wang, L.~Zhang, X.~Tong, L.~Zhang, Z.~Zhang, H.~Liu, X.~Xing, and P.~T.
  Mathiopoulos, ``A three-layered graph-based learning approach for remote
  sensing image retrieval,'' {\em IEEE Transactions on Geoscience and Remote
  Sensing}, vol.~54, no.~10, pp.~6020--6034, 2016.

\bibitem{tong2019exploiting}
X.-Y. Tong, G.-S. Xia, F.~Hu, Y.~Zhong, M.~Datcu, and L.~Zhang, ``Exploiting
  deep features for remote sensing image retrieval: A systematic
  investigation,'' {\em IEEE Transactions on Big Data}, vol.~6, no.~3,
  pp.~507--521, 2019.

\bibitem{martha2011segment}
T.~R. Martha, N.~Kerle, C.~J. Van~Westen, V.~Jetten, and K.~V. Kumar, ``Segment
  optimization and data-driven thresholding for knowledge-based landslide
  detection by object-based image analysis,'' {\em IEEE transactions on
  geoscience and remote sensing}, vol.~49, no.~12, pp.~4928--4943, 2011.

\bibitem{cheng2013automatic}
G.~Cheng, L.~Guo, T.~Zhao, J.~Han, H.~Li, and J.~Fang, ``Automatic landslide
  detection from remote-sensing imagery using a scene classification method
  based on bovw and plsa,'' {\em International Journal of Remote Sensing},
  vol.~34, no.~1, pp.~45--59, 2013.

\bibitem{stumpf2011object}
A.~Stumpf and N.~Kerle, ``Object-oriented mapping of landslides using random
  forests,'' {\em Remote sensing of environment}, vol.~115, no.~10,
  pp.~2564--2577, 2011.

\bibitem{kim2009forest}
M.~Kim, M.~Madden, and T.~A. Warner, ``Forest type mapping using
  object-specific texture measures from multispectral ikonos imagery,'' {\em
  Photogrammetric Engineering \& Remote Sensing}, vol.~75, no.~7, pp.~819--829,
  2009.

\bibitem{capolupo2018novel}
A.~Capolupo, L.~Kooistra, and L.~Boccia, ``A novel approach for detecting
  agricultural terraced landscapes from historical and contemporaneous
  photogrammetric aerial photos,'' {\em International journal of applied earth
  observation and geoinformation}, vol.~73, pp.~800--810, 2018.

\bibitem{mishra2014mapping}
N.~B. Mishra and K.~A. Crews, ``Mapping vegetation morphology types in a dry
  savanna ecosystem: Integrating hierarchical object-based image analysis with
  random forest,'' {\em International Journal of Remote Sensing}, vol.~35,
  no.~3, pp.~1175--1198, 2014.

\bibitem{zhao2015dirichlet}
B.~Zhao, Y.~Zhong, G.-S. Xia, and L.~Zhang, ``Dirichlet-derived multiple topic
  scene classification model for high spatial resolution remote sensing
  imagery,'' {\em IEEE Transactions on Geoscience and Remote Sensing}, vol.~54,
  no.~4, pp.~2108--2123, 2015.

\bibitem{yao2016semantic}
X.~Yao, J.~Han, G.~Cheng, X.~Qian, and L.~Guo, ``Semantic annotation of
  high-resolution satellite images via weakly supervised learning,'' {\em IEEE
  Transactions on Geoscience and Remote Sensing}, vol.~54, no.~6,
  pp.~3660--3671, 2016.

\bibitem{wu2016hierarchical}
H.~Wu, B.~Liu, W.~Su, W.~Zhang, and J.~Sun, ``Hierarchical coding vectors for
  scene level land-use classification,'' {\em Remote Sensing}, vol.~8, no.~5,
  p.~436, 2016.

\bibitem{liu2021multiscale}
R.~Liu, X.~Ning, W.~Cai, and G.~Li, ``Multiscale dense cross-attention
  mechanism with covariance pooling for hyperspectral image scene
  classification,'' {\em Mobile Information Systems}, vol.~2021, 2021.

\bibitem{jiang2019multi}
J.~Jiang, F.~Liu, Y.~Xu, H.~Huang, {\em et~al.}, ``Multi-spectral rgb-nir image
  classification using double-channel cnn,'' {\em IEEE Access}, vol.~7,
  pp.~20607--20613, 2019.

\bibitem{chen2019new}
L.~Chen, X.~Cui, Z.~Li, Z.~Yuan, J.~Xing, X.~Xing, and Z.~Jia, ``A new deep
  learning algorithm for sar scene classification based on spatial statistical
  modeling and features re-calibration,'' {\em Sensors}, vol.~19, no.~11,
  p.~2479, 2019.

\bibitem{wang2018scene}
Q.~Wang, S.~Liu, J.~Chanussot, and X.~Li, ``Scene classification with recurrent
  attention of vhr remote sensing images,'' {\em IEEE Transactions on
  Geoscience and Remote Sensing}, vol.~57, no.~2, pp.~1155--1167, 2018.

\bibitem{zhu2015land}
Y.~Zhu and S.~Newsam, ``Land use classification using convolutional neural
  networks applied to ground-level images,'' in {\em Proceedings of the 23rd
  SIGSPATIAL International Conference on Advances in Geographic Information
  Systems}, pp.~1--4, 2015.

\bibitem{antoniou2016investigating}
V.~Antoniou, C.~C. Fonte, L.~See, J.~Estima, J.~J. Arsanjani, F.~Lupia,
  M.~Minghini, G.~Foody, and S.~Fritz, ``Investigating the feasibility of
  geo-tagged photographs as sources of land cover input data,'' {\em ISPRS
  International Journal of Geo-Information}, vol.~5, no.~5, p.~64, 2016.

\bibitem{pei2014new}
T.~Pei, S.~Sobolevsky, C.~Ratti, S.-L. Shaw, T.~Li, and C.~Zhou, ``A new
  insight into land use classification based on aggregated mobile phone data,''
  {\em International Journal of Geographical Information Science}, vol.~28,
  no.~9, pp.~1988--2007, 2014.

\bibitem{zhang2017parcel}
W.~Zhang, W.~Li, C.~Zhang, D.~M. Hanink, X.~Li, and W.~Wang, ``Parcel-based
  urban land use classification in megacity using airborne lidar, high
  resolution orthoimagery, and google street view,'' {\em Computers,
  Environment and Urban Systems}, vol.~64, pp.~215--228, 2017.

\bibitem{kang2018building}
J.~Kang, M.~K{\"o}rner, Y.~Wang, H.~Taubenb{\"o}ck, and X.~X. Zhu, ``Building
  instance classification using street view images,'' {\em ISPRS journal of
  photogrammetry and remote sensing}, vol.~145, pp.~44--59, 2018.

\bibitem{hu2016mapping}
T.~Hu, J.~Yang, X.~Li, and P.~Gong, ``Mapping urban land use by using landsat
  images and open social data,'' {\em Remote Sensing}, vol.~8, no.~2, p.~151,
  2016.

\bibitem{liu2017classifying}
X.~Liu, J.~He, Y.~Yao, J.~Zhang, H.~Liang, H.~Wang, and Y.~Hong, ``Classifying
  urban land use by integrating remote sensing and social media data,'' {\em
  International Journal of Geographical Information Science}, vol.~31, no.~8,
  pp.~1675--1696, 2017.

\bibitem{jia2018urban}
Y.~Jia, Y.~Ge, F.~Ling, X.~Guo, J.~Wang, L.~Wang, Y.~Chen, and X.~Li, ``Urban
  land use mapping by combining remote sensing imagery and mobile phone
  positioning data,'' {\em Remote Sensing}, vol.~10, no.~3, p.~446, 2018.

\bibitem{tu2018portraying}
W.~Tu, Z.~Hu, L.~Li, J.~Cao, J.~Jiang, Q.~Li, and Q.~Li, ``Portraying urban
  functional zones by coupling remote sensing imagery and human sensing data,''
  {\em Remote sensing}, vol.~10, no.~1, p.~141, 2018.

\bibitem{hong2019cospace}
D.~Hong, N.~Yokoya, J.~Chanussot, and X.~X. Zhu, ``Cospace: Common subspace
  learning from hyperspectral-multispectral correspondences,'' {\em IEEE
  Transactions on Geoscience and Remote Sensing}, vol.~57, no.~7,
  pp.~4349--4359, 2019.

\bibitem{li2017building}
X.~Li, C.~Zhang, and W.~Li, ``Building block level urban land-use information
  retrieval based on google street view images,'' {\em GIScience \& Remote
  Sensing}, vol.~54, no.~6, pp.~819--835, 2017.

\bibitem{cao2018integrating}
R.~Cao, J.~Zhu, W.~Tu, Q.~Li, J.~Cao, B.~Liu, Q.~Zhang, and G.~Qiu,
  ``Integrating aerial and street view images for urban land use
  classification,'' {\em Remote Sensing}, vol.~10, no.~10, p.~1553, 2018.

\bibitem{srivastava2019understanding}
S.~Srivastava, J.~E. Vargas-Munoz, and D.~Tuia, ``Understanding urban landuse
  from the above and ground perspectives: A deep learning, multimodal
  solution,'' {\em Remote sensing of environment}, vol.~228, pp.~129--143,
  2019.

\bibitem{machado2020airound}
G.~Machado, E.~Ferreira, K.~Nogueira, H.~Oliveira, M.~Brito, P.~H.~T. Gama, and
  J.~A. dos Santos, ``Airound and cv-brct: Novel multiview datasets for scene
  classification,'' {\em IEEE Journal of Selected Topics in Applied Earth
  Observations and Remote Sensing}, vol.~14, pp.~488--503, 2020.

\bibitem{deng2018semi}
Z.~Deng, H.~Sun, and S.~Zhou, ``Semi-supervised ground-to-aerial adaptation
  with heterogeneous features learning for scene classification,'' {\em ISPRS
  International Journal of Geo-Information}, vol.~7, no.~5, p.~182, 2018.

\bibitem{han2021trusted}
Z.~Han, C.~Zhang, H.~Fu, and J.~T. Zhou, ``Trusted multi-view classification,''
  {\em arXiv preprint arXiv:2102.02051}, 2021.

\bibitem{sensoy2018evidential}
M.~Sensoy, L.~Kaplan, and M.~Kandemir, ``Evidential deep learning to quantify
  classification uncertainty,'' {\em Advances in neural information processing
  systems}, vol.~31, 2018.

\bibitem{baltruvsaitis2018multimodal}
T.~Baltru{\v{s}}aitis, C.~Ahuja, and L.-P. Morency, ``Multimodal machine
  learning: A survey and taxonomy,'' {\em IEEE transactions on pattern analysis
  and machine intelligence}, vol.~41, no.~2, pp.~423--443, 2018.

\bibitem{ramachandram2017deep}
D.~Ramachandram and G.~W. Taylor, ``Deep multimodal learning: A survey on
  recent advances and trends,'' {\em IEEE signal processing magazine}, vol.~34,
  no.~6, pp.~96--108, 2017.

\bibitem{tu2004fast}
T.-M. Tu, P.~S. Huang, C.-L. Hung, and C.-P. Chang, ``A fast
  intensity-hue-saturation fusion technique with spectral adjustment for ikonos
  imagery,'' {\em IEEE Geoscience and Remote sensing letters}, vol.~1, no.~4,
  pp.~309--312, 2004.

\bibitem{amolins2007wavelet}
K.~Amolins, Y.~Zhang, and P.~Dare, ``Wavelet based image fusion techniques—an
  introduction, review and comparison,'' {\em ISPRS Journal of photogrammetry
  and Remote Sensing}, vol.~62, no.~4, pp.~249--263, 2007.

\bibitem{nunez1999multiresolution}
J.~Nunez, X.~Otazu, O.~Fors, A.~Prades, V.~Pala, and R.~Arbiol,
  ``Multiresolution-based image fusion with additive wavelet decomposition,''
  {\em IEEE Transactions on Geoscience and Remote sensing}, vol.~37, no.~3,
  pp.~1204--1211, 1999.

\bibitem{liu2007bidimensional}
Z.~Liu, P.~Song, J.~Zhang, and J.~Wang, ``Bidimensional empirical mode
  decomposition for the fusion of multispectral and panchromatic images,'' {\em
  International Journal of Remote Sensing}, vol.~28, no.~18, pp.~4081--4093,
  2007.

\bibitem{aiazzi2002context}
B.~Aiazzi, L.~Alparone, S.~Baronti, and A.~Garzelli, ``Context-driven fusion of
  high spatial and spectral resolution images based on oversampled
  multiresolution analysis,'' {\em IEEE Transactions on geoscience and remote
  sensing}, vol.~40, no.~10, pp.~2300--2312, 2002.

\bibitem{zhang2020novel}
R.~Zhang, X.~Tang, S.~You, K.~Duan, H.~Xiang, and H.~Luo, ``A novel
  feature-level fusion framework using optical and sar remote sensing images
  for land use/land cover (lulc) classification in cloudy mountainous area,''
  {\em Applied Sciences}, vol.~10, no.~8, p.~2928, 2020.

\bibitem{wang2022multi}
X.~Wang, Y.~Feng, R.~Song, Z.~Mu, and C.~Song, ``Multi-attentive hierarchical
  dense fusion net for fusion classification of hyperspectral and lidar data,''
  {\em Information Fusion}, vol.~82, pp.~1--18, 2022.

\bibitem{ye2012robust}
G.~Ye, D.~Liu, I.-H. Jhuo, and S.-F. Chang, ``Robust late fusion with rank
  minimization,'' in {\em 2012 IEEE Conference on Computer Vision and Pattern
  Recognition}, pp.~3021--3028, IEEE, 2012.

\bibitem{gunes2005affect}
H.~Gunes and M.~Piccardi, ``Affect recognition from face and body: early fusion
  vs. late fusion,'' in {\em 2005 IEEE international conference on systems, man
  and cybernetics}, vol.~4, pp.~3437--3443, IEEE, 2005.

\bibitem{yu2018aerial}
Y.~Yu and F.~Liu, ``Aerial scene classification via multilevel fusion based on
  deep convolutional neural networks,'' {\em IEEE Geoscience and Remote Sensing
  Letters}, vol.~15, no.~2, pp.~287--291, 2018.

\bibitem{yang2018dropband}
N.~Yang, H.~Tang, H.~Sun, and X.~Yang, ``Dropband: A simple and effective
  method for promoting the scene classification accuracy of convolutional
  neural networks for vhr remote sensing imagery,'' {\em IEEE Geoscience and
  Remote Sensing Letters}, vol.~15, no.~2, pp.~257--261, 2018.

\bibitem{zhang2015scene}
F.~Zhang, B.~Du, and L.~Zhang, ``Scene classification via a gradient boosting
  random convolutional network framework,'' {\em IEEE Transactions on
  Geoscience and Remote Sensing}, vol.~54, no.~3, pp.~1793--1802, 2015.

\bibitem{lin2013cross}
T.-Y. Lin, S.~Belongie, and J.~Hays, ``Cross-view image geolocalization,'' in
  {\em Proceedings of the IEEE Conference on Computer Vision and Pattern
  Recognition}, pp.~891--898, 2013.

\bibitem{lin2015learning}
T.-Y. Lin, Y.~Cui, S.~Belongie, and J.~Hays, ``Learning deep representations
  for ground-to-aerial geolocalization,'' in {\em Proceedings of the IEEE
  Conference on Computer Vision and Pattern Recognition}, pp.~5007--5015, 2015.

\bibitem{workman2017unified}
S.~Workman, M.~Zhai, D.~J. Crandall, and N.~Jacobs, ``A unified model for near
  and remote sensing,'' in {\em Proceedings of the IEEE International
  Conference on Computer Vision}, pp.~2688--2697, 2017.

\bibitem{badrinarayanan2017segnet}
V.~Badrinarayanan, A.~Kendall, and R.~Cipolla, ``Segnet: A deep convolutional
  encoder-decoder architecture for image segmentation,'' {\em IEEE Transactions
  on Pattern Analysis and Machine Intelligence}, vol.~39, no.~12,
  pp.~2481--2495, 2017.

\bibitem{hoffmann2019model}
E.~J. Hoffmann, Y.~Wang, M.~Werner, J.~Kang, and X.~X. Zhu, ``Model fusion for
  building type classification from aerial and street view images,'' {\em
  Remote Sensing}, vol.~11, no.~11, p.~1259, 2019.

\bibitem{srivastava2018multilabel}
S.~Srivastava, J.~E. Vargas-Mu{\~n}oz, D.~Swinkels, and D.~Tuia, ``Multilabel
  building functions classification from ground pictures using convolutional
  neural networks,'' in {\em Proceedings of the 2nd ACM SIGSPATIAL
  International Workshop on AI for Geographic Knowledge Discovery}, pp.~43--46,
  2018.

\bibitem{jiang2018trust}
H.~Jiang, B.~Kim, M.~Guan, and M.~Gupta, ``To trust or not to trust a
  classifier,'' {\em Advances in neural information processing systems},
  vol.~31, 2018.

\bibitem{beyer1999nearest}
K.~Beyer, J.~Goldstein, R.~Ramakrishnan, and U.~Shaft, ``When is “nearest
  neighbor” meaningful?,'' in {\em International conference on database
  theory}, pp.~217--235, Springer, 1999.

\bibitem{blundell2015weight}
C.~Blundell, J.~Cornebise, K.~Kavukcuoglu, and D.~Wierstra, ``Weight
  uncertainty in neural network,'' in {\em International conference on machine
  learning}, pp.~1613--1622, PMLR, 2015.

\bibitem{corbiere2019addressing}
C.~Corbi{\`e}re, N.~Thome, A.~Bar-Hen, M.~Cord, and P.~P{\'e}rez, ``Addressing
  failure prediction by learning model confidence,'' {\em Advances in Neural
  Information Processing Systems}, vol.~32, 2019.

\bibitem{lakshminarayanan2017simple}
B.~Lakshminarayanan, A.~Pritzel, and C.~Blundell, ``Simple and scalable
  predictive uncertainty estimation using deep ensembles,'' {\em Advances in
  neural information processing systems}, vol.~30, 2017.

\bibitem{van2020uncertainty}
J.~Van~Amersfoort, L.~Smith, Y.~W. Teh, and Y.~Gal, ``Uncertainty estimation
  using a single deep deterministic neural network,'' in {\em International
  conference on machine learning}, pp.~9690--9700, PMLR, 2020.

\bibitem{damianou2013deep}
A.~Damianou and N.~D. Lawrence, ``Deep gaussian processes,'' in {\em Artificial
  intelligence and statistics}, pp.~207--215, PMLR, 2013.

\bibitem{heo2018uncertainty}
J.~Heo, H.~B. Lee, S.~Kim, J.~Lee, K.~J. Kim, E.~Yang, and S.~J. Hwang,
  ``Uncertainty-aware attention for reliable interpretation and prediction,''
  {\em Advances in neural information processing systems}, vol.~31, 2018.

\bibitem{moon2020confidence}
J.~Moon, J.~Kim, Y.~Shin, and S.~Hwang, ``Confidence-aware learning for deep
  neural networks,'' in {\em international conference on machine learning},
  pp.~7034--7044, PMLR, 2020.

\bibitem{neumann2018relaxed}
L.~Neumann, A.~Zisserman, and A.~Vedaldi, ``Relaxed softmax: Efficient
  confidence auto-calibration for safe pedestrian detection,'' 2018.

\bibitem{yager2008classic}
R.~R. Yager and L.~Liu, {\em Classic works of the Dempster-Shafer theory of
  belief functions}, vol.~219.
\newblock Springer, 2008.

\bibitem{josang2016subjective}
A.~J{\o}sang, {\em Subjective logic}, vol.~3.
\newblock Springer, 2016.

\bibitem{lin2016dirichlet}
J.~Lin, ``On the dirichlet distribution,'' {\em Department of Mathematics and
  Statistics, Queens University}, 2016.

\bibitem{simonyan2014very}
K.~Simonyan and A.~Zisserman, ``Very deep convolutional networks for
  large-scale image recognition,'' {\em arXiv preprint arXiv:1409.1556}, 2014.

\bibitem{krizhevsky2017imagenet}
A.~Krizhevsky, I.~Sutskever, and G.~E. Hinton, ``Imagenet classification with
  deep convolutional neural networks,'' {\em Communications of the ACM},
  vol.~60, no.~6, pp.~84--90, 2017.

\bibitem{he2016deep}
K.~He, X.~Zhang, S.~Ren, and J.~Sun, ``Deep residual learning for image
  recognition,'' in {\em Proceedings of the IEEE conference on computer vision
  and pattern recognition}, pp.~770--778, 2016.

\bibitem{vo2016localizing}
N.~N. Vo and J.~Hays, ``Localizing and orienting street views using overhead
  imagery,'' in {\em European conference on computer vision}, pp.~494--509,
  Springer, 2016.

\bibitem{geng2022multi}
W.~Geng, W.~Zhou, and S.~Jin, ``Multi-view urban scene classification with a
  complementary-information learning model,'' {\em Photogrammetric Engineering
  \& Remote Sensing}, vol.~88, no.~1, pp.~65--72, 2022.

\bibitem{szegedy2016rethinking}
C.~Szegedy, V.~Vanhoucke, S.~Ioffe, J.~Shlens, and Z.~Wojna, ``Rethinking the
  inception architecture for computer vision,'' in {\em Proceedings of the IEEE
  conference on computer vision and pattern recognition}, pp.~2818--2826, 2016.

\bibitem{huang2017densely}
G.~Huang, Z.~Liu, L.~Van Der~Maaten, and K.~Q. Weinberger, ``Densely connected
  convolutional networks,'' in {\em Proceedings of the IEEE conference on
  computer vision and pattern recognition}, pp.~4700--4708, 2017.

\bibitem{liang2022advances}
W.~Liang, G.~A. Tadesse, D.~Ho, L.~Fei-Fei, M.~Zaharia, C.~Zhang, and J.~Zou,
  ``Advances, challenges and opportunities in creating data for trustworthy
  ai,'' {\em Nature Machine Intelligence}, vol.~4, no.~8, pp.~669--677, 2022.

\end{thebibliography}
\end{document}